%% file: main.tex
\documentclass[]{bytedance}
\usepackage[toc,page,header]{appendix}

\usepackage{minitoc}
\usepackage{amsfonts}
\usepackage{amssymb}
\usepackage{tabularx}
\usepackage{listings}
\usepackage{xcolor}
\usepackage{cancel}

\usepackage{tabulary,multirow,xspace}
\usepackage{fixmath,mathtools,nicefrac,mmstyle}
\usepackage{subcaption}
\usepackage{caption}
\usepackage{wrapfig} 
\usepackage[misc]{ifsym} 
\usepackage{colortbl}

\usepackage{wrapfig}
\usepackage{multicol}
\usepackage[most]{tcolorbox}
\usepackage{pifont}

\usepackage{graphicx}
\usepackage[export]{adjustbox}
\usepackage{booktabs}
\usepackage{afterpage}

\definecolor{codegreen}{rgb}{0,0.6,0}
\definecolor{codegray}{rgb}{0.5,0.5,0.5}
\definecolor{codepurple}{rgb}{0.58,0,0.82}
\definecolor{backcolour}{rgb}{0.95,0.95,0.92}
\definecolor{boxblue}{RGB}{57,89,163}
\definecolor{boxbluebg}{RGB}{230,237,250} 
\definecolor{myblue}{RGB}{210, 225, 255}

\lstdefinestyle{mystyle}{
    backgroundcolor=\color{backcolour},   
    commentstyle=\color{codegreen},
    keywordstyle=\color{magenta},
    numberstyle=\tiny\color{codegray},
    stringstyle=\color{codepurple},
    basicstyle=\ttfamily\footnotesize,
    breakatwhitespace=false,         
    breaklines=true,                 
    captionpos=b,                    
    keepspaces=true,                 
    numbers=none,                    
    numbersep=5pt,                  
    showspaces=false,                
    showstringspaces=false,
    showtabs=false,                  
    tabsize=2
}
\definecolor{mygray1}{gray}{.95}
\definecolor{mygray2}{gray}{.9}
\definecolor{mygray3}{gray}{.95}
\usepackage{pifont}

\newlength\savewidth
\newcolumntype{x}[1]{>{\centering\arraybackslash}p{#1pt}}

\newcommand{\app}{\raise.17ex\hbox{$\scriptstyle\sim$}}

\renewcommand{\emph}[1]{\textit{#1}}

\usepackage{xcolor}
\input{macros}

\usepackage{listings}

\definecolor{commentgreen}{rgb}{0.1, 0.4, 0.1}
\definecolor{keywordblue}{rgb}{0.1, 0.1, 0.7}
\definecolor{stringred}{rgb}{0.7, 0.1, 0.1}

\lstdefinestyle{mystyle}{
    commentstyle=\color{commentgreen},
    keywordstyle=\color{keywordblue},   
    stringstyle=\color{stringred},
    basicstyle=\ttfamily\scriptsize, 
    breaklines=true,
    keepspaces=true,
    showstringspaces=false,
    frame=none,                     
    language=Python, 
}

\title{
    DreamOmni3: Scribble-based Editing and Generation
}

\author[]{Bin Xia$^{1,2}$}
\author[]{Bohao Peng$^{1}$}
\author[]{JiyangLiu$^{2}$}
\author[]{Sitong Wu$^{1}$} 
\author[]{Jingyao Li$^{1}$}
\author[]{Junjia Huang$^{2}$}
\author[]{\\Xu Zhao$^{2}$}
\author[]{Yitong Wang$^{2}$}
\author[]{Ruihang Chu$^{1}$}
\author[]{Bei Yu$^{1}$}
\author[]{Jiaya Jia $^3$}

\affiliation[]{CUHK, HKUST, ByteDance}

\abstract{
Recently unified generation and editing models have achieved remarkable success with their impressive performance. These models rely mainly on text prompts for instruction-based editing and generation, but language often fails to capture users’ intended edit locations and fine-grained visual details. To this end, we propose two tasks: scribble-based editing and generation, which enable more flexible creation on a graphical user interface (GUI) combining user textual, images, and freehand scribbles. We introduce DreamOmni3, tackling two challenges: data creation and framework design. Our data synthesis pipeline includes two parts: scribble-based editing and generation. For scribble-based editing, we define four tasks: scribble and instruction-based editing, scribble and multimodal instruction-based editing, image fusion, and doodle editing. Based on DreamOmni2’s dataset, we extract editable regions and overlay hand-drawn boxes, circles, doodles, or cropped images to construct training data. For scribble-based generation, we define three tasks: scribble and instruction-based generation, scribble and multimodal instruction-based generation, and doodle generation, following similar data creation pipelines. For the framework, instead of using binary masks, which struggle with complex edits involving multiple scribbles, images, and text instructions, we propose a joint input scheme that feeds both the original and scribbled source images into the model, using different colors to distinguish regions and simplify processing. By applying the same index and position encodings to both images, the model can precisely localize scribbled regions while maintaining accurate editing. Furthermore,
we introduce joint training with the VLM and our generation/editing model to better understand the user's abstract scribbles. Finally, we establish comprehensive DreamOmni3 benchmarks for these tasks to promote further research. Experimental results show that DreamOmni3 achieves outstanding performance.
}

\date {November 20, 2025}

\checkdata[Project Page]{\url{https://zj-binxia.github.io/DreamOmni3-projectpage/}}

\begin{document}
\maketitle

\begin{figure}
\centering
    \includegraphics[width=\linewidth]{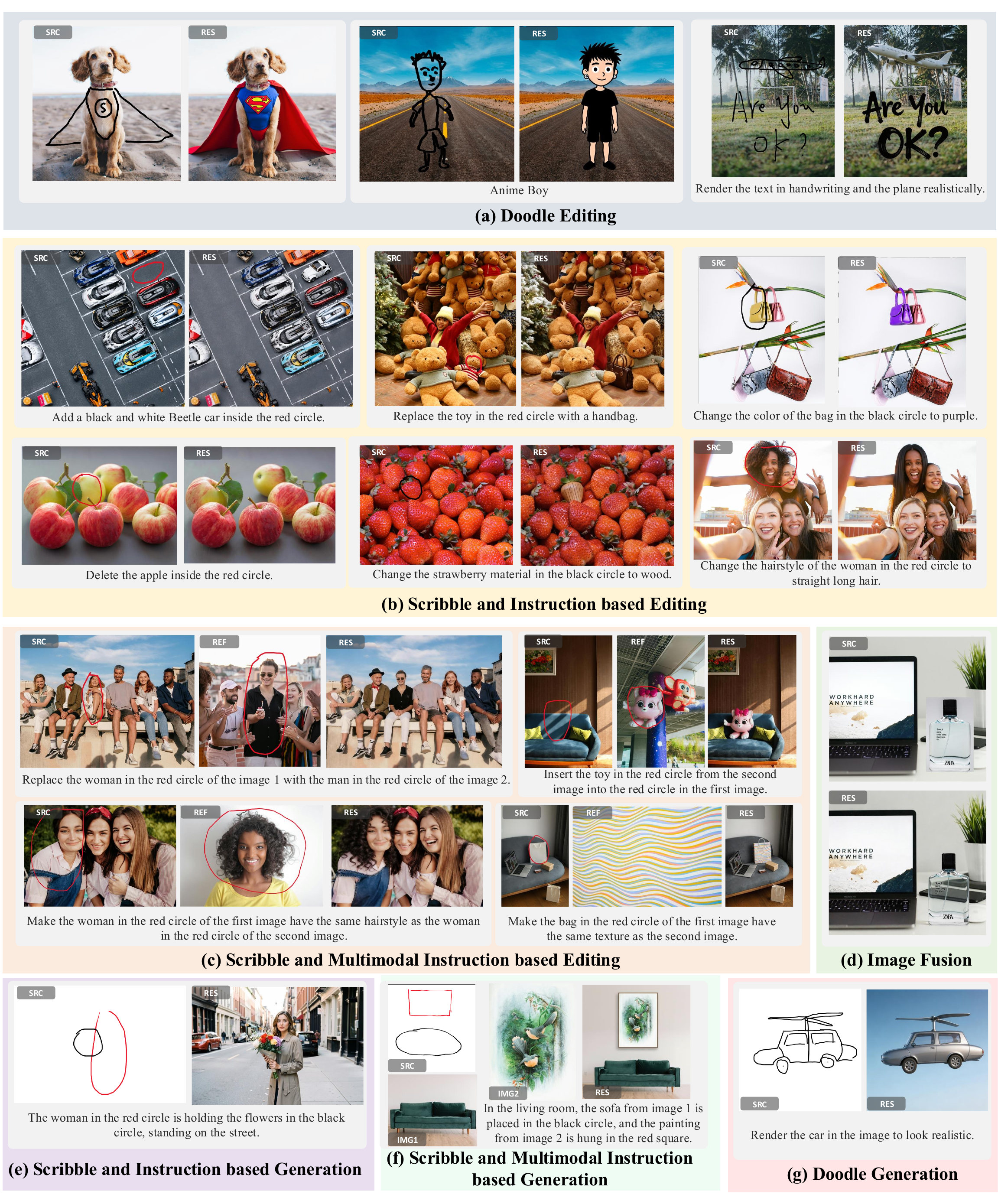}
    \caption{DreamOmni3 gallery showcasing scribble-based editing and generation, including doodle editing/generation, scribble and instruction-based editing/generation, scribble and multimodal-instruction-based editing/generation, and image fusion.
    }
    \label{fig:summary}
\end{figure}

\section{Introduction}
The recent success of unified generation and editing models~\cite{nanobanana,gpt4o,bagel} can be attributed to three key factors:
\textbf{(1)} Unified training for generation and editing not only improves performance on existing tasks through mutual enhancement but also gives rise to a wide range of new editing and generation capabilities.
\textbf{(2)} It significantly lowers the user barrier, enabling all functions within a single model without the need to choose specialized ones.
\textbf{(3)} They exhibit strong multimodal understanding, responding to real-world visuals, marking progress toward world models and AGI.

The current unified generation and editing models focus on generation and editing based on images and text as instructions. However, this approach often falls short of meeting users' interactive needs. For example, certain editing or generation positions in the image are difficult to describe with language, some objects may be hard for users to identify by name, or there may be multiple identical objects that are hard to distinguish. In these cases, manual annotation is necessary. Furthermore, users may wish to make more flexible and creative additions, deletions, or modifications to the content in the image, such as through drawing. In these scenarios, it's essential for the model to not only understand images and language but also understand the editing and generation intentions behind the user's manual scribbles.

To create a more intelligent and comprehensive unified creation tool, we present DreamOmni3, a model that integrates our proposed scribble-based editing and generation with the existing unified editing and generation framework. First, we define the tasks of scribble-based editing and generation to guide our data creation. Specifically, as shown in Fig.~\ref{fig:summary}, we categorize scribble-based editing into four types: scribble and instruction-based editing, scribble and multimodal instruction-based editing, image fusion, and doodle editing. Additionally, we categorize scribble-based generation into three types: scribble and instruction-based generation, scribble and multimodal instruction-based generation, and doodle generation.

To train DreamOmni3, the main challenge is the lack of training data. To address this, we introduce a comprehensive data synthesis pipeline for scribble-based editing and generation. \textbf{Scribble-based editing data creation}: \textbf{(1)} For multimodal instruction-based editing, we use the Referseg service~\cite{lisa} on the DreamOmni2 dataset to obtain the coordinates and size of the edited objects in both the source and reference images, marking the edited positions with hand-drawn circles or boxes. \textbf{(2)} For instruction-based editing, we use the same data as in step (1) but without the reference image. \textbf{(3)} For image fusion, we extract the edited objects from the reference image and paste them onto the corresponding position on the source image. \textbf{(4)} For doodle editing, we crop the edited objects from the target image, generate sketches, and place them back into the source image. \textbf{Scribble-based generation data creation}: \textbf{(1)} For multimodal instruction-based generation, we use Referseg to locate the edited objects in the image and mark the circles or boxes on the blank canvas. \textbf{(2)} For instruction-based generation, we remove the reference image from the data created in step (1). \textbf{(3)} For doodle generation, we extract the edited objects from the target image, generate sketches, and place them on a blank canvas at the same position.

For the framework, we chose to input the source image with scribbles rather than using a binary mask for two reasons (Fig.~\ref{fig:train_data}): \textbf{(1)} It avoids the complexity and computational cost of binary masks when editing multiple areas. Scribbles, distinguished by color, are easier to describe with instructions. \textbf{(2)} It is compatible with existing unified editing and generation models, which require RGB inputs, not binary masks.

Based on the above analysis, we propose the DreamOmni3 framework. Specifically, we propose a joint input scheme for editing by feeding both the original source image and the scribbled source image. This allows us to retain both the scribbled information and the pixels covered by the scribbles. To better align the source image with its scribbled version, both share the same index and position encodings. For the reference image, to distinguish it from the source images and avoid pixel confusion, we adopt the same position shift and index encoding scheme as in DreamOmni2 (Fig.~\ref{fig:method}~(c)). Additionally, we propose a joint training scheme for the generation/editing model and VLM. Since user scribbles are often diverse and VLM possesses strong reasoning capabilities, we train it on our scribbles dataset to recognize and provide contextual explanations of the user's scribbles. This aids the generation/editing model in better understanding the user's intent and producing more accurate results.
Furthermore, for these tasks, we construct a DreamOmni3 benchmark using real-world image data, enabling a more accurate evaluation of the model’s generalization and real-world performance. Our main contributions are fourfold:

 \begin{itemize}
 \item We introduce two highly useful tasks for unified generation and editing models: scribble-based editing and scribble-based generation. These tasks can be combined with text and image instructions, enhancing the creative usability of unified models and providing clear definitions for targeted optimization and future research.
\item  We propose a data synthesis pipeline to create a high-quality, comprehensive training dataset for scribble-based editing and generation.
\item We present DreamOmni3, a framework that supports text, image, and scribble inputs with complex logic. It takes both the source text, image and scribble as input, maintaining editing consistency while accurately interpreting the scribble’s intent. We design position and index encoding schemes to differentiate between the scribble and source image, ensuring compatibility with existing unified generation and editing architectures. Additionally, we propose joint training of the generation/editing model and VLM to enhance scribble understanding.
\item For these tasks, we build the DreamOmni3 benchmark using real-world image data. Experiments demonstrate the effectiveness of DreamOmni3 in real scenarios.
\end{itemize}

\section{Related Work}
\label{sec:formatting}

\subsection{Mask-based Generation} 
Image generation has evolved from text-conditioned synthesis~\cite{LDM,dalle2,Imagen,sdxl,sd3,pixart,hunyuan-dit,dreamomni,sindiffusion,zhang2025aligning} toward more controllable paradigms that incorporate richer visual and multimodal conditions. Subject-driven image generation~\cite{textual-inversion,dreambooth,blipdiffusion,xverse,dreamo,iclora,ip-adapter,anydoor, CustomDiffusion, SuTI, Instantid,invadapter,migc++,ominicontrol}, further enables models to preserve the identity or appearance of a given subject while synthesizing it under new contexts. More recently, multimodal instruction-based generation~\cite{dreamomni2} has extended this flexibility by jointly leveraging textual instructions and visual references, allowing users to specify both semantic intent and visual content in a more expressive manner.

Unlike the aforementioned generation paradigms, mask-based generation introduces explicit masks to spatially control the content generated within specified regions. It can be broadly categorized into three types.
For mask-guided text-to-image generation, Multi-diffusion~\cite{msdiffusion} first explored a training-free approach for multi-region generation on text-to-image models, but the results were unstable. GLIGEN~\cite{gligen} improved this by introducing an attention module and encoding bounding boxes through Fourier embeddings. Later works~\cite{migc,instancediffusion,eligen} further optimized attention mechanisms and refined the granularity of input information.
For mask-guided subject-driven generation, MS-diffusion~\cite{ms-diffusion} employs a grounding resampler to associate visual information with specific entities and spatial constraints.
Nevertheless, existing mask-based generation frameworks are often overly complex in both input format and inference process. In contrast, DreamOmni3 allows users to paint a mask and combine it with image and language instructions to generate more complex outputs. Its unified design aligns with existing frameworks, bridging mask-based generation and image editing within a single system, greatly enhancing usability.

\subsection{Mask-based Editing} 

Image editing can be broadly divided into training-free~\cite{sdedit,prompttoprompt,NulltextInversion,pnp,masactrl,diffedit,flowedit,infedit,ledits} and learning-based paradigms~\cite{instructp2p, bagel,emu-edit,llmga,puma,unireal,magicbrush,dreamomni,omniedit,seed-edit,dreamve,step1x,mgie,ultraedit,anyedit}. Training-free methods directly manipulate pretrained generative models without task-specific training, while instruction-based methods learn editing capabilities from large-scale image-instruction editing datasets.

Furthermore, mask-based editing refers to editing operations performed on selected regions or items of an image. It can generally be divided into three categories:
\textbf{(1)} Image inpainting~\cite{diffir,xia2024diffi2i,Paintbyexample,dreamomni,llmga} is a common technique where users paint over the area they wish to modify. The model then regenerates that region based on the instruction while preserving the rest of the image.
\textbf{(2)} Since traditional inpainting completely regenerates the masked region and cannot preserve its original structure or color, MagicQuill~\cite{magicquill} introduces auxiliary inputs such as edge maps or low-resolution images to help maintain contour and color consistency.
\textbf{(3)} Some approaches~\cite{smartmask,mao2025ace++,insertanything} inject compressed object IDs from a reference image into the masked region to insert specific objects.

Compared to the above mask-based editing methods, DreamOmni3 offers several advantages:
\textbf{(a)} Traditional inpainting datasets are created by simply masking parts of an image, which prevents the model from reasoning about environmental effects such as lighting or shadow changes (the third row of Fig.~\ref{fig:qual-edit} ). Our dataset, built through instruction-based editing, allows the model to respond more coherently to contextual changes.
\textbf{(b)} Inpainting often suffers from inaccurate masks — regions may be over- or under-edited. Even MagicQuill, though guided by edge maps, still requires precise manual masking. In contrast, our instruction-based approach is far more robust: users can roughly circle a region, and the model accurately understands both the intent and the editing scope, providing a much more user-friendly experience.
\textbf{(c)} The unified generation and editing model is the way forward. However, previous methods relied on inconsistent input formats, making them hard to unify under a single framework. DreamOmni3 establishes a unified architecture that simplifies use and integration.
\textbf{(d)} Benefiting from this unified design, DreamOmni3 not only extends existing editing capabilities but also introduces new ones. For example, in image fusion, prior methods~\cite{mao2025ace++} could only merge a single reference image into a masked area, whereas our model supports multimodal inputs (multiple images, text, scribbles, and more) for richer and more precise edits. Moreover, DreamOmni3 bridges the gap between image generation and editing, enabling interactive experiences like direct sketching on tablets.

\section{Methodology}

\begin{figure*}[t]
\centering
    \includegraphics[width=1\linewidth]{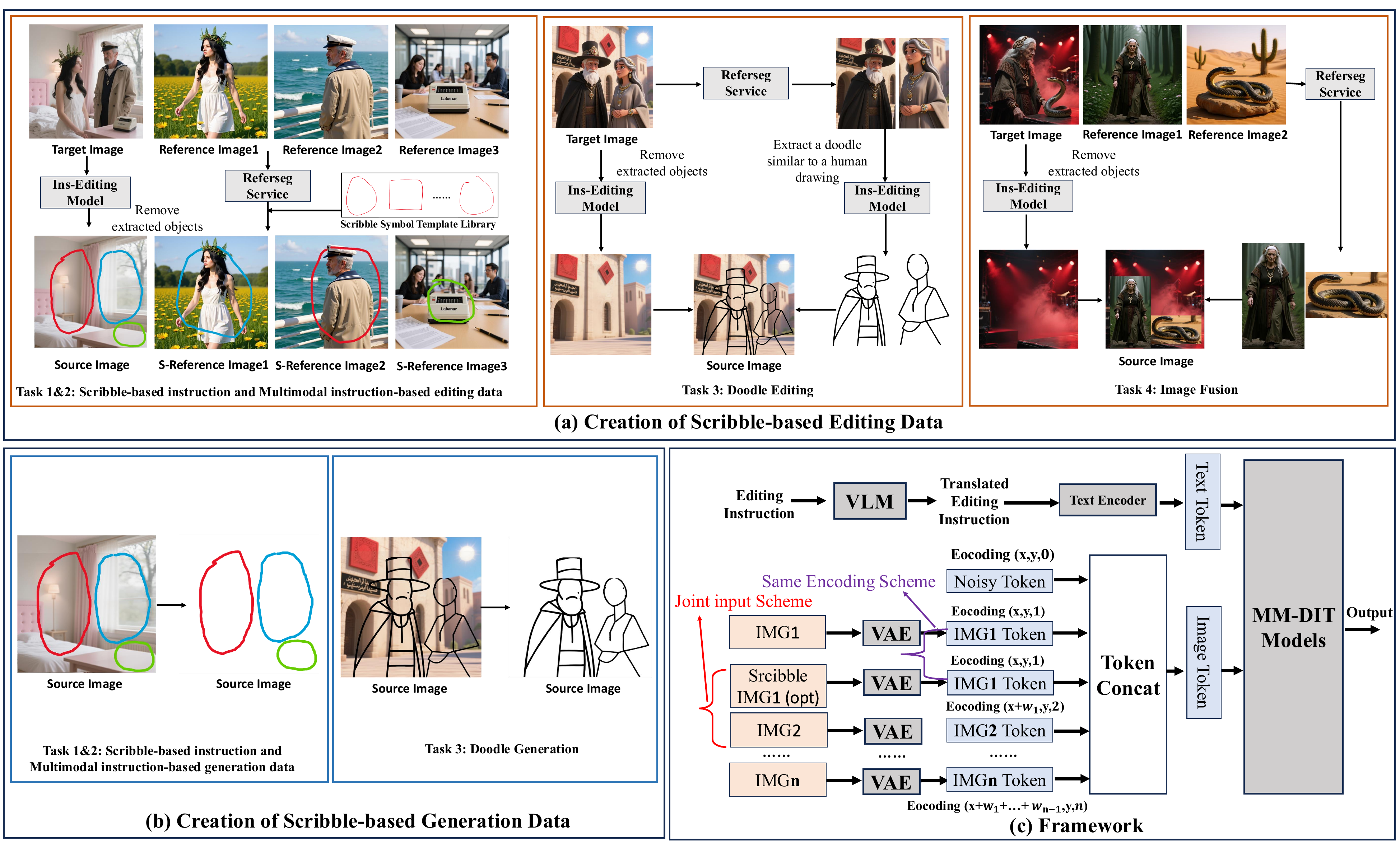}
    \captionof{figure}{The overview of DreamOmni3's training data construction and framework. 
 The overview of DreamOmni3's training data construction and framework:
\textbf{(a)} We create scribble-based editing training data. For scribble and multimodal instruction-based editing, we use Referseg~\cite{lisa} to locate edit objects and paste corresponding scribbles onto the source and reference images to create training pairs. For scribble and instruction-based editing, we omit the reference image. For doodle editing, we use a dedicated model to convert the edit objects into abstract sketches and paste them back into the source image. For image fusion, we crop objects from the reference image and paste them into the corresponding position on the source image to build training pairs.
\textbf{(b)} Scribble-based generation training data is created similarly to editing, except the source image is replaced with a blank white canvas.
\textbf{(c)} DreamOmni3 builds on the framework of DreamOmni2~\cite{dreamomni2}, introducing a joint input scheme for scribble inputs. We also apply the same encoding scheme to both the source and scribbled images, ensuring better pixel alignment and perfect compatibility with previous image and language instruction editing. }
    \label{fig:method}
\end{figure*}

\begin{figure}
\centering
    \captionsetup{type=figure}
    \includegraphics[width=1\linewidth]{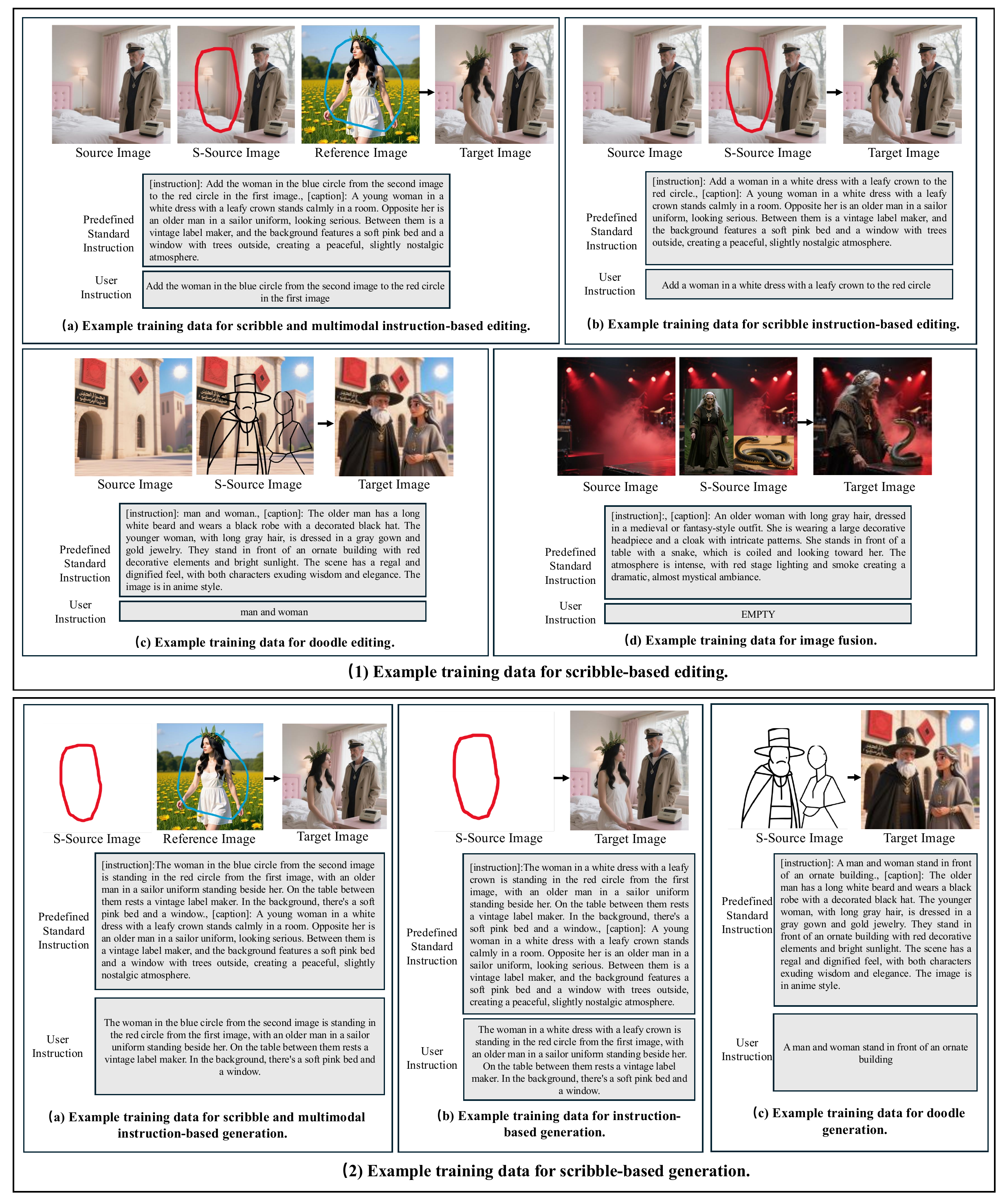}
    \captionof{figure}{Example training data for scribble-based editing and generation. }
    \label{fig:train_data}
\end{figure}

\subsection{Synthetic Data}
The biggest challenge in scribble-based editing and generation is the lack of data. We need to construct a dataset that incorporates language, images, and scribbles as instructions, and develop the ability to combine these three instruction types for complex editing tasks, enabling smarter editing tools. We found that  DreamOmni2 effectively unified language and image instructions and introduced the multimodal instruction editing and generation task, along with corresponding training data. Therefore, we directly use DreamOmni2's multimodal instruction editing and generation data as the base, and extend it to create a dataset that includes scribbles as instructions.

As shown in Fig.~\ref{fig:method} (a), we divide scribble-based editing into four tasks:
\textbf{(1)} Scribble and multimodal instruction-based editing: We use Referseg~\cite{lisa} service to locate the editing objects in both the reference and target images. Since users often draw imperfect shapes, we manually created $300$ different squares and circles for a scribble symbol template library. After determining the object’s position in the reference image, we randomly select and resize a symbol from the library, apply data augmentation to the symbol by varying its line thickness, color, size, and rotation, and then paste it onto the reference image to create a scribble-based reference image (S-Reference Image). For the source image, we modify the target image by either removing or altering the object, then add the corresponding scribble symbol after data augmentation, resulting in the scribble-based source image (S-source image). This process generates target images based on either the reference image (or S-reference image) and the source image. The reference image can include or exclude scribbles depending on the instruction. Our editing types encompass a wide range of both concrete objects and abstract attributes, as demonstrated in DreamOmni2~\cite{dreamomni2}.
\textbf{(2)} Scribble-based instruction editing: This task requires no special data preparation. We simply remove the reference image from (1) and add an object description to the instruction.
\textbf{(3)} Doodle editing: We use Referseg to locate the editing object and the instruction-based editing model to turn it into a simple abstract doodle. We avoid using Canny edge detection due to users' imperfect drawings, which require aesthetic corrections from the model. Instead, we use the GPT-Image-1 model, which is better suited for generating doodles with non-strict pixel consistency.
\textbf{(4)} Image fusion: users can extract objects from one image and insert them into the target image. We first remove the target object's corresponding section using the instruction-based editing model. Next, we use Referseg to locate and crop the object from the reference image, then paste it into the target image after resizing it to fit, obtaining the source image.

As shown in Fig.~\ref{fig:method} (b), we divide scribble-based generation into three tasks: \textbf{(1)} For scribble-based multimodal instruction generation, the method is similar to that for scribble-based multimodal instruction editing, with the key difference being the placement of scribble symbols on a white canvas to create the source image. This enables the model to refer to attributions or objects of the reference image and generate them at corresponding symbol positions. \textbf{(2)} For scribble-based instruction generation, we remove the reference image from step (1) and add descriptions of the removed objects in the instructions. \textbf{(3)} For doodle generation, the method is similar to doodle editing, with the final sketch placed on a white canvas to allow the model to generate the corresponding objects and background based on the sketch and instructions.

Our data is created based on DreamOmni2's multi-reference image generation/editing training datasets. Our scribble-based editing dataset consists of several types of data, with representative examples shown in Fig.~\ref{fig:train_data} (1).
 The scribble-based multimodal instruction editing includes 32K training samples, while scribble-based instruction editing has about 14K training samples. The image fusion dataset consists of 16K training samples, and the doodle editing dataset contains 8K training samples. Notably, both scribble-based multimodal instruction editing and scribble-based instruction editing cover a wide range of editing categories. These include edits related to abstract properties, such as design style, color scheme, and hairstyle, as well as edits of concrete objects, such as various objects, people, and animals, with the ability to add, remove, or modify them. In contrast, image fusion and doodle editing focus on the task of adding concrete objects into the image.

The scribble-based generation dataset contains several types of training data, with representative examples shown in Fig.~\ref{fig:train_data}(2).
Specifically, scribble-based multimodal instruction generation contains 29K training samples, scribble-based instruction generation contains 10K training samples, and doodle generation contains 8K training samples.
 Both scribble-based multimodal instruction generation and scribble-based instruction generation involve generating concrete objects as well as referencing abstract attributions. However, doodle generation is focused on generating concrete objects.

\begin{figure}
\centering
    \captionsetup{type=figure}
    \includegraphics[width=0.7\linewidth]{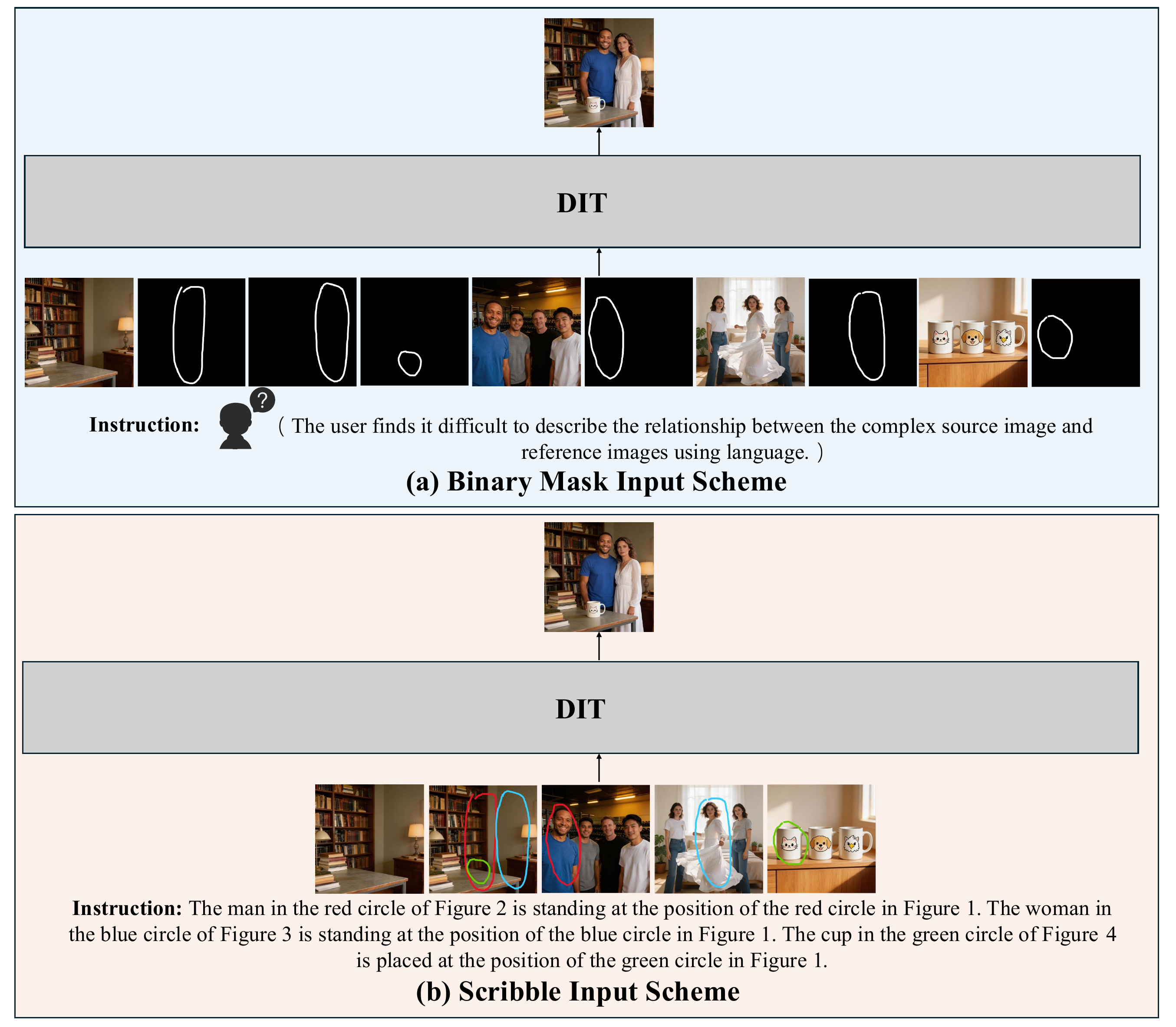}
    \captionof{figure}{Comparison of the traditional binary mask input scheme and our scribble input scheme. }
    \label{fig:scheme}
\end{figure}

\subsection{Framework and Training}

The current unified generation and editing models primarily focus on instruction-based editing and subject-driven generation. Recently, DreamOmni2 extended this model to multi-reference generation and editing. However, the input format for doodle instructions still requires exploration. In DreamOmni3, we considered two input schemes: one using binary masks, similar to inpainting, and another using a joint input of the source image and the source image with doodles. Since doodles inevitably alter parts of the source image, but editing requires consistency in the non-edited areas, we need to input doodle information while preserving the source image details.

As shown in Fig.~\ref{fig:scheme}, there are two possible approaches: the traditional binary mask input scheme (inpainting) and our proposed scribble input scheme. Our approach offers several advantages over binary mask scheme:
\textbf{(1)} Scribble inputs make it easier to express complex relationships between the reference and source images, as users can distinguish different scribbles by referring to their colors in natural language instructions. In contrast, binary masks are limited to black and white, making complex task specification more difficult.
\textbf{(2)} Scribble inputs are more computationally efficient, as multiple scribbles can be drawn on a single image and distinguished by different colors. In contrast, binary masks require a separate image for each mask, increasing the input and computational overhead as the number of masks grows.
\textbf{(3)} The scribble input scheme integrates naturally with existing unified generation and editing models, which are typically trained on RGB images, whereas binary masks cannot fully leverage their pretrained visual capabilities.

Additionally, we found that user scribbles are often messy, requiring the model to have strong reasoning abilities to interpret their intent. Thus, we introduce a VLM with enhanced reasoning capabilities. However, since existing VLMs are not specifically trained to predict the use of scribbles, they need to be trained on our scribble dataset to accurately infer user intent and output a predefined annotation format. To this end, we propose joint training of the VLM and generation/editing models, allowing the VLM to interpret complex text, image, and scribble instructions and improve the generation/editing model’s understanding of user intent.

Our framework design is shown in Fig.~\ref{fig:method} (c), building upon the DreamOmni2 framework with additional adaptation for scribble instruction input. Our joint input scheme is optional. Specifically, when the source image in the editing task contains scribbles, we input both the source image and the scribbled source image into the MM-DIT models. However, if the reference images contain scribbles, we do not use the joint input scheme, as there is no need to maintain pixel-level consistency in the non-edited areas of the reference image, and adding an extra input would unnecessarily increase computational cost. For scribble-based generation tasks, we also avoid using the joint input scheme since there is no need for pixel-level preservation.
Using the joint input scheme does introduce two challenges: \textbf{(1)} it adds an extra image, which affects the indexing of subsequent input images and could potentially confuse users, and \textbf{(2)} the model must correctly map pixel relationships between the source image and the doodle-modified source image. To address these, we use the same index encoding and position encoding scheme for both the source image and the scribbled source image. Experiments show that this encoding effectively resolves these issues, seamlessly integrating doodle editing capabilities into the existing unified generation and editing model framework.

During training, we fine-tune Qwen2.5-VL~\cite{qwen2-vl} 7B to learn the predefined standard output format, with a learning rate of $1 \times 10^{-5}$, using approximately 10 A100 hours. For more details, please refer to the appendix.
We then train the editing and generation models using LoRA with a rank of $256$. Notably, by leveraging LoRA for training, we retain the original instruction-editing capabilities of Kontext. When a user inputs an image with scribbles into the model, our LoRA is activated, seamlessly integrating multimodal instruction-based editing and generation into the unified model. Additionally, since the multi-reference generation and editing capabilities in the previous DreamOmni2 model were trained separately with two LoRAs, our generation and editing models are also trained with separate LoRAs to ensure compatibility. The training process took approximately $400$ A100 hours.

\subsection{Benchmark and Evaluation}
\label{sec:evaluation}
We propose scribble-based editing and generation that integrates language, images, and scribble instructions. Currently, there is no dedicated benchmark to evaluate these tasks. To foster the development of this valuable direction, we introduce the DreamOmni3 benchmark. Our benchmark is comprehensive, consisting of real images to accurately assess model performance in real-world scenarios. The test cases cover a wide range of editing and generation tasks, including the four editing tasks and three generation tasks proposed in our paper. The editing categories are also diverse, encompassing both abstract property edits and concrete object edits. Furthermore, the scribbles in our DreamOmni3 benchmark are manually drawn by different users to better evaluate the model under real-world user interactions. More details about the DreamOmni3 benchmark can be found in the appendix.

Traditional metrics, such as DINO~\cite{dino} and CLIP~\cite{clip}, are not sufficient to accurately evaluate our complex and diverse scribble-based editing and generation tasks. Recent works like Step1x-Edit~\cite{step1x} and Kontext~\cite{kontext} rely on VLMs and human evaluations for instruction editing. Given the increased complexity of our Scribble-based Editing and Generation tasks, evaluation must be based on VLMs. We propose a comprehensive set of evaluation standards for these new tasks, focusing on three key aspects: \textbf{(1)} accuracy in following the instructions in the generated edits, \textbf{(2)} consistency in human appearance, objects, and abstract attributes, \textbf{(3)} avoidance of severe visual artifacts, and \textbf{(4)} alignment of the generated or edited content with the specified doodle regions. Only when all these criteria are met do we consider the editing or generation task successful, ensuring relevance to real-world use cases. We apply the same standards for human evaluations, and our results show that VLM-based assessments align closely with human evaluations. The system prompt used for VLM evaluation can be found in the appendix.

\begin{table*}[t]
  \centering
  \caption{Quantitative comparison of scribble-based editing and generation. We use Gemini~\cite{gemini} and Doubao~\cite{doubao} to evaluate the success editing ratio of different models on scribble-based editing and generation, respectively. In addition, ``Human'' refers to professional engineers assessing the editing success rates of all models.}
    \begin{tabular}{l|ccc|ccc}
    \toprule[0.2em]
    \multirow{2}[2]{*}{\textbf{Method}} & \multicolumn{3}{c|}{\textbf{Scribble-based Editing}} & \multicolumn{3}{c}{\textbf{Scribble-based Generation}} \\
          & \textbf{Gemini$\uparrow$} & \textbf{Doubao$\uparrow$} & \textbf{Human$\uparrow$} & \textbf{Gemini$\uparrow$} & \textbf{Doubao$\uparrow$} & \textbf{Human$\uparrow$} \\
    \midrule
    GPT-4o~\cite{gpt4o} & 0.6176  & 0.5441  & 0.5956  & 0.6023  & 0.5114  & 0.4318  \\
    Nano Banana~\cite{nanobanana} & 0.4779  & 0.3897  & 0.4118  & 0.4318  & 0.3977  & 0.2727  \\
    \midrule
    Omnigen2~\cite{omnigen2} & 0.0735  & 0.0588  & 0.0294  & 0.0909  & 0.0795  & 0.0341  \\
    Kontext~\cite{kontext} & 0.0515  & 0.1029  & 0.0147  & 0.0682  & 0.0227  & 0.0909  \\
    Qwen-image-Edit-2509~\cite{qwen-image} & 0.1912  & 0.1691  & 0.1397  & 0.1364  & 0.0568  & 0.0795  \\
    DreamOmni2~\cite{dreamomni2} & 0.1765  & 0.2059  & 0.1471  & 0.1477  & 0.0795  & 0.0568  \\
    \midrule
    DreamOmni3 (Ours) & \textbf{0.5294}  & \textbf{0.4632}  & \textbf{0.5588}  & \textbf{0.5341}  & \textbf{0.4773}  & \textbf{0.5455}  \\
    \bottomrule[0.2em]
    \end{tabular}%
  \label{tab:quan-edit-gen}%
\end{table*}%

\begin{table}[htbp]
  \centering
  \caption{Comparison of consistency under CLIP and DINO.}
    \begin{tabular}{lcc}
    \toprule[0.2em]
    \textbf{Method} & \textbf{CLIP-I$\uparrow$} & \textbf{DINO$\uparrow$} \\
    \midrule
    GPT-4o~\cite{gpt4o} & 0.8368  & 0.7585  \\
    Nano Banana~\cite{nanobanana} & 0.8669  & 0.8186  \\
    \midrule
    Omnigen2~\cite{omnigen2} & 0.8611  & 0.8029  \\
    Kontext~\cite{kontext} & 0.8278  & 0.7443  \\
    Qwen-image-Edit-2509~\cite{qwen-image} & 0.8565  & 0.8155  \\
    DreamOmni2~\cite{dreamomni2} & 0.8076  & 0.7026  \\
    \midrule
    DreamOmni3 (Ours) & \textbf{0.8702}  & \textbf{0.8172}  \\
    \bottomrule[0.2em]
    \end{tabular}%
  \label{tab:traditional}%
\end{table}%

\begin{figure}
\centering
    \includegraphics[width=1\linewidth]{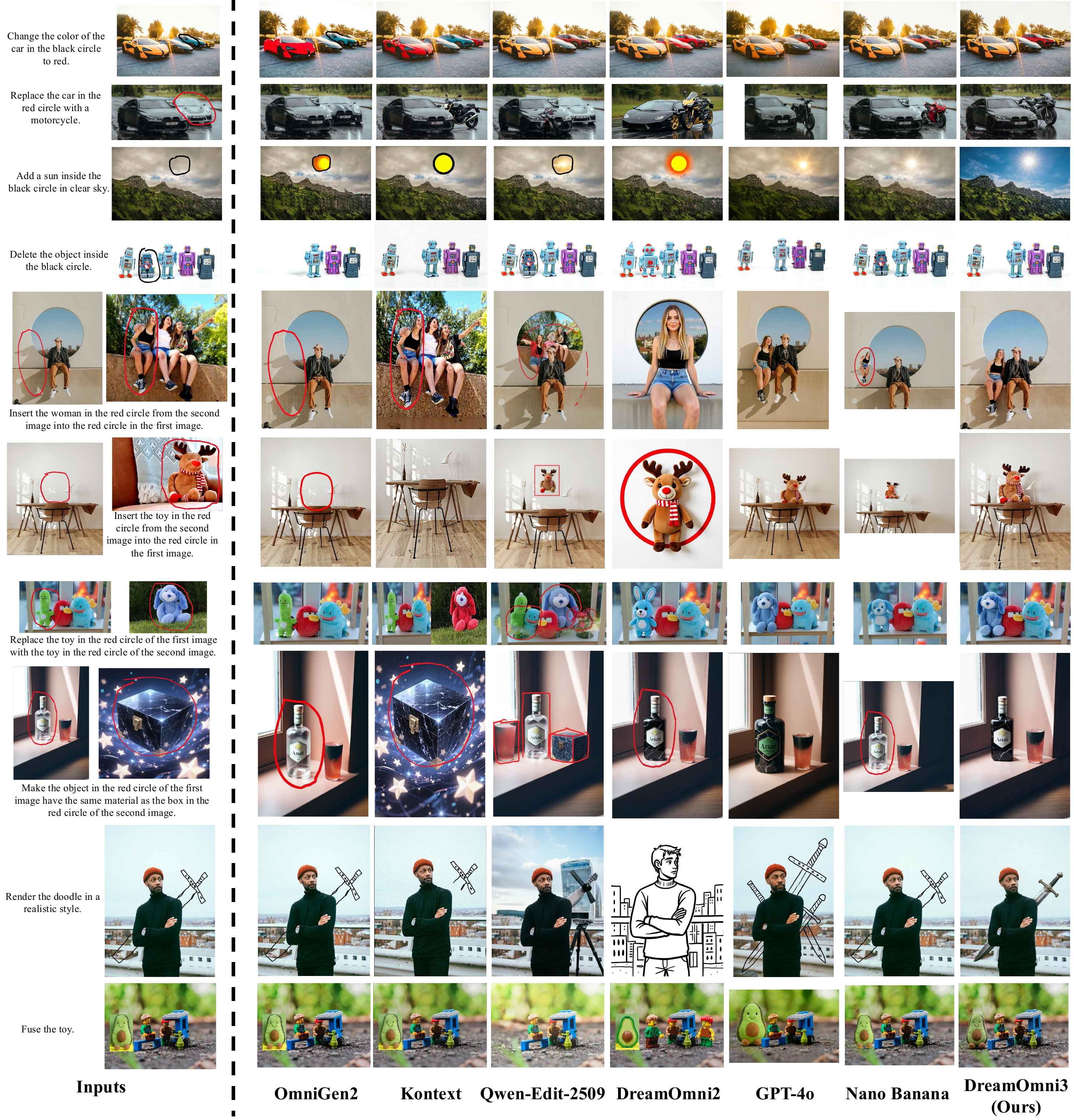}
    \captionof{figure}{Visual comparison of scribble-based editing. Compared to other competitive methods and even closed-source commercial models (GPT-4o and Nano Banana), DreamOmni3 shows better editing results and consistency. }
    \label{fig:qual-edit}
\end{figure}

\section{Experiments}

\textbf{Evaluation on Scribble-based Editing.}
As shown in Tab.~\ref{tab:quan-edit-gen}, we compared several competitive unified generation and editing models, such as Omnigen2~\cite{omnigen2}, Qwen-image-Edit-2509~\cite{qwen-image}, and DreamOmni2~\cite{dreamomni2}. In addition, although Kontext does not natively support multi-image input, we applied a method from Diffusers~\cite{diffusers} that combines multiple images into one for input. We also compared closed-source commercial models, Nano Banana~\cite{nanobanana} and GPT-4o~\cite{gpt4o}. We tested the performance of all models on the DreamOmni3 benchmark, which we constructed using real images for doodle editing, and evaluated the model outputs' pass rates using Gemini 2.5~\cite{gemini} and Doubao 1.6~\cite{doubao}. Additionally, we had several people manually evaluate the results from different models and derive the pass rates. The evaluation standards are outlined in Sec.~\ref{sec:evaluation}. For manual assessment, each case is evaluated by 5 reviewers, and a case is deemed successful if it receives approval from more than 3 evaluators. As can be seen, DreamOmni3 achieved the best results in human evaluations. For Vision Language Model (VLM) evaluations, DreamOmni3 also outperformed open-source models and showed results comparable to commercial models. Notably, GPT-4o frequently exhibits a yellowing issue in the images, and the pixels in the non-edited areas of the output often do not match those in the input image. Meanwhile, Nano Banana showed several issues, such as copy-and-paste effects and incorrect object proportions. These issues are difficult for VLMs to detect accurately.

Furthermore, as shown in Tab.~\ref{tab:traditional}, we evaluate feature-level editing consistency using CLIP-I~\cite{clip} and DINO~\cite{dino} scores. DreamOmni3 outperforms open-source models and approaches closed-source commercial models.

The qualitative results are shown in Fig.~\ref{fig:qual-edit}. We present a range of editing outcomes, including scribble-based instruction edits in the first four rows, scribble-based multimodal instruction edits in rows 5 to 8, and a comparison of doodle editing results in row 9, followed by image fusion results in row 10. As observed, DreamOmni3 generates more precise edits with better consistency.

\begin{figure}
\centering
    \captionsetup{type=figure}
    \includegraphics[width=1\linewidth]{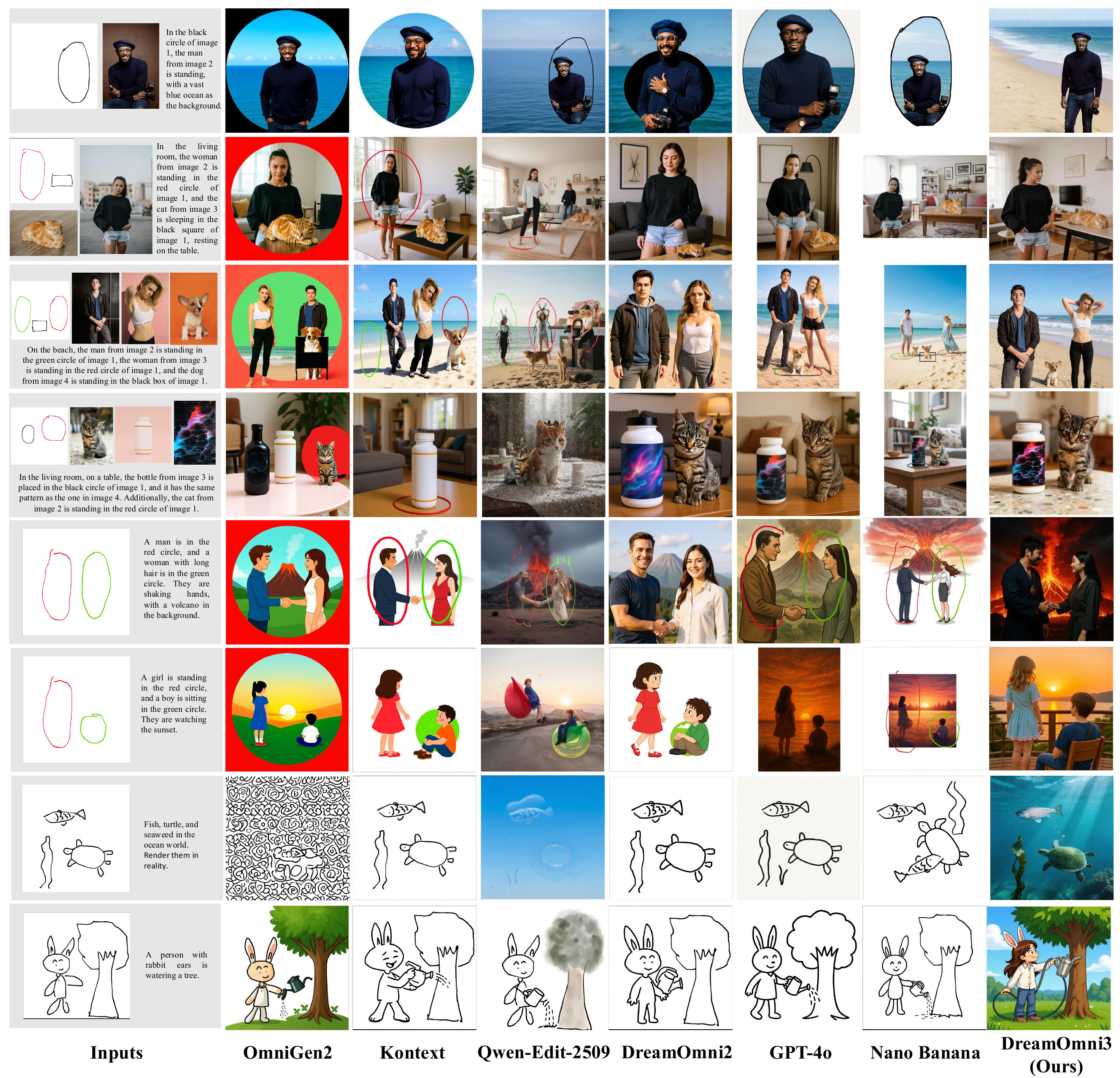}
    \captionof{figure}{Visual comparison of scribble-based generation. DreamOmni3 significantly outperforms current open-source models and achieves generation results comparable to closed-source commercial models (GPT-4o and Nano Banana). }
    \label{fig:qual-gen}
\end{figure}

\noindent\textbf{Evaluation on Scribble-based Generation.} 
As shown in Tab.~\ref{tab:quan-edit-gen}, our method outperforms Nano Banana in both human evaluations and assessments by Doubao and Gemini, with performance similar to GPT-4o. In fact, we found that GPT-4o and Nano Banana are not specifically optimized for scribble-based scenarios, and even when explicitly instructed not to generate scribble marks, these models still often output scribbles along with the generated results. The evaluation standards are outlined in Sec.~\ref{sec:evaluation}, and further details can be found in the appendix. Compared to open-source models, the current state-of-the-art (SOTA) models, DreamOmni2 and Qwen-image-edit-2509, also do not focus on or optimize for these new tasks, resulting in lower success rates on scribble-related tasks. This further demonstrates the effectiveness and necessity of our DreamOmni3 for scribble-based generation.

Qualitative results are shown in Fig.~\ref{fig:qual-gen}. We observe that open-source models often fail to accurately follow user instructions and tend to retain the input scribbles in their outputs. Commercial models, such as Nano Banana and GPT-4o, perform relatively better, although Nano Banana sometimes produces collage-like artifacts.

\noindent\textbf{Agreement of VLM-based and human evaluations. } 
As shown in Tab.~\ref{tab:ins-edit}, we measure the agreement between human and VLM-based evaluations using Cohen's $\kappa$ across different scribble-based editing and generation tasks. Our automated VLM evaluation scheme shows strong agreement with human judgments across all tasks, enabling fair comparison for future work, real-time monitoring of training progress, and its use as a reward model for reinforcement learning.

\begin{table}[t]
  \centering
  \caption{Cohen's $\kappa$ Agreement Between Human and VLM Evaluations Across Different Tasks.}
  \resizebox{\linewidth}{!}{
    \begin{tabular}{c|cccc|ccc}
      \toprule[0.2em]
      \multirow[c]{2}{*}[-1.3em]{\bfseries Tasks}
      & \multicolumn{4}{c|}{\textbf{Scribble-based Editing}}
      & \multicolumn{3}{c}{\textbf{Scribble-based Generation}} \\
      \cmidrule{2-8}
      & \textbf{\shortstack{Doodle \\ Editing}}
      & \textbf{\shortstack{Scribble \\ and \\ Ins-based \\ Editing}}
      & \textbf{\shortstack{Scribble \\ and Multimodal \\ Ins-based \\ Editing}}
      & \textbf{\shortstack{Image \\ Fusion}}
      & \textbf{\shortstack{Doodle \\ Generation}}
      & \textbf{\shortstack{Scribble \\ and \\ Ins-based \\ Generation}}
      & \textbf{\shortstack{Scribble \\ and Multimodal \\ Ins-based \\ Generation}} \\
      \midrule
      \shortstack{Cohen's $\kappa$ \\ (Gemini) $\uparrow$}
      & 0.4792 & 0.7872 & 0.6257 & 0.4595
      & 0.5385 & 0.6479 & 0.5970 \\
      \midrule
      \shortstack{Cohen's $\kappa$ \\(Doubao) $\uparrow$}
      & 0.5192 & 0.7902 & 0.5714 & 0.4521
      & 0.5385 & 0.7235 & 0.5844 \\
      \bottomrule[0.2em]
    \end{tabular}%
  }
  \label{tab:ins-edit}
\end{table}

\noindent\textbf{Joint Input and Joint Training.}
As shown in Tab.~\ref{tab:joint-input}, we compare the impact of training with our custom scribbled dataset versus using joint inputs of source images and scribbled source images, along with the joint training of VLM, on model performance. In Scheme 1, we use the base model, Kontext, which only takes the scribbled source image as input.
In Scheme 2, we train the model on our dataset, but input only the scribbled source image, without using the trained VLM to assist in understanding the scribble. 
In Scheme 3, we combine the source image and the scribbled source image as joint input, but without training on our dataset or using the trained VLM.
In Scheme 4, we use the base model Kontext (not trained on our dataset) and input only the scribbled source image, but with a trained VLM to assist in understanding.
Following the same pattern, in Scheme 8, we use joint input, train the generation/editing models on our dataset, and leverage the trained VLM to assist in understanding the scribbles. 

Based on experiments, we can draw three key conclusions.
 \textbf{(1)} Comparing Scheme 2 with Scheme 1, we observe that our dataset significantly improves the model's scribble-based editing and generation capabilities. \textbf{(2)} Comparing Scheme 6 with Scheme 8, joint input significantly boosts editing performance, while the improvement in generation tasks is less pronounced. This is because scribbles can obscure edited regions of the source image, so joint input ensures the model can see the original image information, improving editing consistency. In contrast, generation tasks don't require high pixel-level consistency, making the improvement less significant. Based on these results, we only use joint input for editing tasks when the source image contains scribbles. For other reference images or generation tasks with scribbles, we input the scribbled image directly. \textbf{(3)} Comparing Scheme 7 and Scheme 8, we find that the trained VLM effectively helps the generation/editing models understand the user's scribble intent, resulting in more accurate outputs.

\begin{table}[t]
  \centering
  \caption{Joint input scheme and joint training scheme for scribble-based generation and editing.}
    \begin{tabular}{c|c|c|c|cc}
    \toprule[0.2em]
    \textbf{Method} & \textbf{\shortstack{Generation or Editing \\Model Training}} & \textbf{Joint Input} & \textbf{Joint Training} & \multicolumn{1}{l}{\textbf{Editing}} & \multicolumn{1}{l}{\textbf{Generation}} \\
    \midrule
    Scheme 1 & \ding{55}     & \ding{55}     & \ding{55}     & 0.0882  & 0.0341  \\
    Scheme 2 & \ding{51}     & \ding{55}     & \ding{55}     & 0.2868   & 0.3409   \\
    Scheme 3 & \ding{55}     & \ding{51}     & \ding{55}     & 0.0956  & 0.0341  \\
    Scheme 4 & \ding{55}     & \ding{55}     & \ding{51}     & 0.0882  & 0.0455  \\
    Scheme 5  & \ding{55}     & \ding{51}     & \ding{51}     & 0.1176  & 0.0455  \\
    Scheme 6 & \ding{51}     & \ding{55}     & \ding{51}     & 0.3897   & 0.4432   \\
    Scheme 7 & \ding{51}     & \ding{51}     & \ding{55}     & 0.3456   & 0.3636   \\
    Scheme 8 (Ours)  & \ding{51}     & \ding{51}     & \ding{51}     & \textbf{0.4632}  & \textbf{0.4773}  \\
    \bottomrule[0.2em]
    \end{tabular}%
  \label{tab:joint-input}%
\end{table}%

\noindent\textbf{Index and Position Encoding.}
As shown in Tab.~\ref{tab:index}, we compare different encoding schemes for the source image and the scribbled source image as joint inputs. The results show that using the same index encoding and position encoding for both inputs yields the best performance. We believe this is due to two main reasons: \textbf{(1)} Matching the position and index encodings allows for better pixel-level alignment between the two images, enabling the model to more accurately locate the scribbles and preserve pixel-level information, leading to more consistent and precise editing. \textbf{(2)} Using the same encodings means that subsequent reference images do not need to modify their encodings, maintaining consistency with the original training format and allowing the model to better leverage its pre-trained editing capabilities for more accurate results.

\begin{table}[t]
  \centering
  \caption{Different encoding schemes for scribble inputs.}
    \begin{tabular}{c|c|c|cc}
    \toprule[0.2em]
    \textbf{Method} & \textbf{\shortstack{Same Index\\ Encoding }} & \textbf{\shortstack{Same Position \\Encoding}} & \textbf{Editing} & \textbf{Generation} \\
    \midrule
    Scheme 1 & \ding{55}     &\ding{55}     & 0.3824  & 0.3068  \\
    Scheme 2 & \ding{55}     & \ding{51}     & 0.4118  & 0.4318  \\
    Scheme 3 & \ding{51}     & \ding{55}     & 0.4338  & 0.3636  \\
    Scheme 4 (Ours) & \ding{51}     & \ding{51}     & \textbf{0.4632}  & \textbf{0.4773}  \\
    \bottomrule[0.2em]
    \end{tabular}%
   \label{tab:index}%
\end{table}%

\noindent\textbf{Fine-Grained Evaluation on Scribble-Based Editing.}
As shown in Tab.~\ref{tab:fine-edit}, we report DreamOmni3's performance across five evaluation metrics on four scribble-based editing tasks. These evaluation metrics include Consistency of Non-Edited Regions measures the proportion of cases where non-edited areas remain unchanged between the source image and result; Instruction Accuracy, which evaluates how well text and scribble instructions are followed; Visual Quality, which assesses the naturalness of the result; Output Scribble, which measures the proportion of results without scribbles; and Consistency of Edited Elements, which evaluates the appearance consistency of edited elements. For more details, please refer to the appendix. We can draw five conclusions: \textbf{(1)} Among the four tasks, Scribble and Multimodal Instruction-based Editing has the lowest overall success rate, indicating that it is the most challenging and require more training data to improve model performance.
\textbf{(2)} For Doodle Editing, the main bottleneck lies in the naturalness of the generated results.
\textbf{(3)} For Scribble and Instruction-based Editing, the key issue is unresponsiveness to instructions.
\textbf{(4)} In Scribble and Multimodal Instruction-based Editing, failure mainly occurs due to the inconsistent appearance of reference objects or humans.
\textbf{(5)} For Image Fusion, the main issue is the poor fusion of the collage image.

\begin{table}[t]
  \centering
  \caption{Fine-Grained DreamOmni3 Evaluation on Scribble-Based Editing. }
  \resizebox{1\linewidth}{!}{
    \begin{tabular}{c|ccccc|c}
    \toprule[0.2em]
    \textbf{\shortstack{Scribble-based \\Editing Tasks}} & \textbf{\shortstack{Consistency of \\Non-Edited Regions}} & \textbf{\shortstack{Instruction\\ Accuracy}} & \textbf{\shortstack{Visual \\Quality }} & \textbf{\shortstack{Output \\ Scribbles}} & \textbf{\shortstack{Consistency of \\Edited Elements}} & \textbf{Overall} \\
    \midrule
    Doodle Editing & 0.7500  & 0.6000  & 0.5500  & 0.8500  & 0.7500  & 0.4500  \\
    \midrule
    \shortstack{Scribble and \\Instruction-based Editing} & 0.7083  & 0.5625  & 0.9167  & 0.7292  & 0.8333  & 0.5208  \\
    \midrule
    \shortstack{Scribble and Multimodal\\Instruction-based Editing} & 0.6042  & 0.4792  & 0.8125  & 0.6458  & 0.4167  & 0.3333  \\
    \midrule
    Image Fusion & 1.0000  & 0.8500  & 0.7500  & /     & 0.9500  & 0.6500  \\
    \bottomrule[0.2em]
    \end{tabular}%
    }
  \label{tab:fine-edit}%
\end{table}%

\noindent\textbf{Fine-Grained Evaluation on Scribble-Based Generation.}
As shown in Tab.~\ref{tab:fine-gen}, we report DreamOmni3's performance across four evaluation metrics on three scribble-based generation tasks. These metrics are: Consistency, which measures the ratio of cases where the result aligns with the reference image or doodle; Instruction Accuracy, which evaluates how well the result follows the text instructions; Visual Quality, which assesses the naturalness of the generated result; and Output Scribble, which measures the proportion of results without scribbles. We can draw four conclusions:
\textbf{(1)} Scribble and Multimodal Instruction-based Generation has the lowest overall success rate among the three tasks, making it the most challenging.
\textbf{(2)} For Doodle Generation, the main difficulty is improving the naturalness and aesthetics of the output.
\textbf{(3)} For Scribble and Instruction-based Generation, the key challenge is unresponsiveness to instructions.
\textbf{(4)} For Scribble and Multimodal Instruction-based Generation, the main issue is maintaining consistency in the appearance of people and objects in the reference image.

\begin{table}[t]
  \centering
  \caption{Fine-Grained DreamOmni3 Evaluation on Scribble-Based Generation.}
    \begin{tabular}{c|cccc|c}
    \toprule[0.2em]
    \textbf{\shortstack{Scribble-based \\Generation Tasks}} & \textbf{Consistency}  & \textbf{\shortstack{Instruction\\ Accuracy}} & \textbf{\shortstack{Visual \\Quality }} & \textbf{\shortstack{Output \\ Scribbles}} & \textbf{Overall} \\
    \midrule
    Doodle Generation & 0.7778  & 0.7222  & 0.6111  & 0.7222  & 0.5556  \\
    \midrule
    \shortstack{Scribble and \\Instruction-based Generation} & /     & 0.6333  & 0.8667  & 0.7667  & 0.5667  \\
    \midrule
    \shortstack{Scribble and Multimodal\\Instruction-based Generation} & 0.4750  & 0.5500  & 0.7750  & 0.6500  & 0.3750  \\
    
    \bottomrule[0.2em]
    \end{tabular}%
  \label{tab:fine-gen}%
\end{table}%

\section{Conclusion}

The current unified generation and editing model performs image edits based on text instructions. However, language struggles to accurately describe edit locations and capture all the user's intended details. To enhance this, we propose two tasks: scribble-based editing and generation, allowing users to simply use a brush in the GUI to make edits. This method can combine language, image, and scribble instructions, offering more flexibility.
Building on this, we introduce DreamOmni3, addressing the challenge of limited training data. Using DreamOmni2 data, we developed a data creation scheme based on Referseg to generate high-quality, accurate datasets that integrate scribbles, text, and image instructions.
We also tackled the model framework issue, as binary masks are inadequate for complex real-world needs. When multiple masks are present, they are hard to distinguish and describe with language. To solve this, we propose a scribble-based approach, where different masks are easily differentiated by brush color, allowing for the handling of any number of masks.
Since scribbles may obscure some image details, we introduce a joint input scheme, feeding both the original and scribbled images into the model. We further optimized this by using the same index and position encoding to preserve the details while maintaining accurate correspondence to reference images. We have also introduced joint training of VLM and generation/editing models to improve the model's understanding of user text, images, and irregular scribbles, enabling more accurate results.
In addition, we developed a DreamOmni3 benchmark and evaluation framework that combines scribbles, text, and image input instructions, fostering the advancement of this new field.

\clearpage

\bibliographystyle{plainnat}
\bibliography{main}

\clearpage
\appendix
\section{The Overview of the Appendix}

\textbf{(1)} We provide more training details of DreamOmni3 (Sec.~\ref{sec:train}).

\textbf{(2)} We provide the system prompt for model evaluation and additional evaluation details (Sec.~\ref{sec:modelevaluation}).

\textbf{(3)} We provide more details on the DreamOmni3 benchmark (Sec.~\ref{sec:benchmark}).

\textbf{(4)} We present a large number of DreamOmni3 visual results on Scribble-based Editing (Sec.~\ref{sec:more_editing}).

\textbf{(5)} We present a large number of DreamOmni3 visual results on Scribble-based Generation (Sec.~\ref{sec:more_gen}).

\section{Training of DreamOmni3}
\label{sec:train}
For the training of DreamOmni3, we have followed the joint training scheme of VLM and Kontext from DreamOmni2~\cite{dreamomni2}. In the first phase, we train the VLM to translate user instructions into predefined standard instructions. In the second phase, we use images and predefined standard instructions as inputs to Kontext~\cite{kontext}, training the model to generate the target image.
mage.

During training, we fine-tune Qwen2.5-VL~\cite{qwen2-vl} 7B to learn the predefined standard output format, with a learning rate of $1 \times 10^{-5}$, using approximately 10 A100 hours. Both DreamOmni3 scribble-based editing and generation LoRA are trained on a batch size of $16$ and a learning rate of $5 \times 10^{-6}$, consuming about 400 A100 hours.

\section{Model Evaluation}
\label{sec:modelevaluation}

As shown in Tabs.~\ref{tab:eva_edit1}, \ref{tab:eva_edit2}, \ref{tab:eva_edit3}, and~\ref{tab:eva_edit4}, we provide the system prompts used on Doubao~\cite{doubao} and Gemini~\cite{gemini} in the paper for evaluating scribble and multimodal instruction-based editing,  scribble instruction-based editing, doodle editing, and image fusion, respectively.

Furthermore, as shown in Tabs.~\ref{tab:eva_gen1}, \ref{tab:eva_gen2}, and~\ref{tab:eva_gen3}, we provide he system prompts used on Doubao~\cite{doubao} and Gemini~\cite{gemini} in the paper for evaluating scribble and multimodal instruction-based generation,  scribble instruction-based generation, and doodle generation, respectively.

For the human evaluation in the Tab.~\textcolor{red}{1} and Tab.~\textcolor{red}{2} of the paper, we use the same evaluation criteria as described in the system prompt above. We will ask five evaluators to assess each scribble-based editing case based on the corresponding criteria above. If any criterion is not met, the case will be marked as a failure; otherwise, it will be considered a successful edit. A case is deemed successful if at least three evaluators judge it as such; otherwise, it will be classified as a failure.

\begin{table*}[h!]\centering

\begin{minipage}{1\linewidth}\vspace{0mm}    \centering
\begin{tcolorbox} 
    \small
You are a professional image editing evaluator. I will provide you with a source image (which may contain scribbles), a reference image (which may contain scribbles), a target image, and an editing instruction. Your task is to evaluate the target image based on the following criteria:

1. Consistency of Non-Edited Regions: Assess whether the regions or objects in the source image that were not edited (as per the instructions) are consistent with the target image. Specifically, compare whether the appearance and position of each object in the non-edited areas of the source and target images remain strictly consistent, and evaluate if there are any significant color discrepancies, such as yellowing, whitening, sharpening, blurring, etc.

2. Instruction Accuracy: First, evaluate whether the model accurately edited the objects described in the scribbles as instructed. For example, if the instruction asks to replace the object in the red circle of the source image with the object in the yellow circle of the reference image, verify if the object in the red circle of the source image was replaced with the object in the yellow circle of the reference image. Besides, check if the editing instruction has been accurately executed. For instance, if the instruction asks to place a woman from the source image onto a sofa in the reference image, verify if the woman in the target image not only looks similar to the reference image but also correctly sits on the sofa as instructed.

3. Consistency of Edited Elements: Evaluate the consistency of the objects or elements in the target image that were edited based on the instruction. Compare the reference image and target image to ensure the requested edits were applied properly.

4. Quality of the Editing Result: Assess the overall quality of the edits in the target image. Compare the edited objects between the reference image and the target image to determine if the edits appear unnatural, such as if the person or object has been copied from the reference image into the target image in a way that doesn't blend well. Also, evaluate whether any parts of the object are missing or if the object doesn't seamlessly integrate with the surrounding environment.

5. Check for scribbles in the Output: Check whether the scribbles inputted by the user are also output in the target image. It is not acceptable for the scribbles to be included in the output image.

Output the result in JSON format as: {"judge": True/False, "reason": "xxx"}. Please analyze the requirements step by step according to the five points. Only return True if all points are satisfied. Please provide the reasoning for each of the five points.

\end{tcolorbox}

\caption{The system prompt for the VLM (Doubao and Gemini) to assess \textbf{scribble and multimodal instruction-based editing}.}
    \label{tab:eva_edit1}
\end{minipage}
\end{table*}

\begin{table*}[h!]\centering

\begin{minipage}{1\linewidth}\vspace{0mm}    \centering
\begin{tcolorbox} 
    \small
You are a professional image editing evaluator. I will provide you with a source image (which may contain scribbles), a target image, and an editing instruction. Your task is to evaluate the target image based on the following criteria:

1. Consistency of Non-Edited Regions: Assess whether the regions or objects in the source image that were not edited (as per the instructions) are consistent with the target image. Specifically, compare whether the appearance and position of each object in the non-edited areas of the source and target images remain strictly consistent, and evaluate if there are any significant color discrepancies, such as yellowing, whitening, sharpening, blurring, etc.

2. Instruction Accuracy: First, evaluate whether the model accurately edited the objects described in the scribbles as instructed. For example, if the instruction asks to replace the object in the red circle of the source image with a dog, verify if the object in the red circle of the source image was replaced with the dog. Besides, check if the editing instruction has been accurately executed. For instance, if the instruction asks to make the woman dance, verify if the woman in the target image is dancing.

3. Quality of the Editing Result: Evaluate whether the edited objects are naturally integrated into the target image. Naturalness is mainly assessed based on lighting, shadows, and the size proportion of the objects. Additionally, check whether the edited objects allow the user to immediately recognize that they are fake or unnatural.

4. Check for scribbles in the Output: Check whether the scribbles inputted by the user are also output in the target image. It is not acceptable for the scribbles to be included in the output image.

Output the result in JSON format as: {"judge": True/False, "reason": "xxx"}. Please analyze the requirements step by step according to the four points. Only return True if all points are satisfied. Please provide the reasoning for each of the four points.

\end{tcolorbox}

\caption{The system prompt for the VLM (Doubao and Gemini) to assess \textbf{scribble instruction-based editing}.}
    \label{tab:eva_edit2}
\end{minipage}
\end{table*}

\begin{table*}[h!]\centering

\begin{minipage}{1\linewidth}\vspace{0mm}    \centering
\begin{tcolorbox} 
    \small
You are a professional image editing evaluator. I will provide you with a source image (which may contain scribbles), a target image, and an instruction (which may or may not be present). Your task is to evaluate the target image based on the following criteria:

1. Consistency of Non-Edited Regions: Assess whether the regions or objects in the source image that were not edited (according to the instructions and scribbles) are consistent with the target image. Specifically, compare whether the appearance and position of each object in the non-edited areas of the source and target images remain strictly consistent, and evaluate if there are any significant color discrepancies, such as yellowing, whitening, sharpening, blurring, etc.

2. Scribbles and Instruction-based Editing Accuracy: You need to determine whether the model has edited the position of the user's scribbles in the target image. The edited result should resemble the scribble but be transformed into a more aesthetically pleasing actual object, rather than just the scribble lines. Additionally, if the user provided instructions, evaluate whether the generated result meets the requirements of the instructions.

3. Quality of the Editing Result: Evaluate whether the edited objects are naturally integrated into the target image. Naturalness is primarily assessed based on lighting, shadows, and the size proportion of the objects. Additionally, check whether the edited objects allow the user to immediately recognize that they are fake or unnatural.

4. Check for Scribbles in the Output: Check whether the scribbles input by the user are also output in the target image. It is not acceptable for the scribbles to be included in the output image.

Output the result in JSON format as: {"judge": True/False, "reason": "xxx"}. Please analyze the requirements step by step according to the four points. Only return True if all points are satisfied. Please provide the reasoning for each of the four points.

\end{tcolorbox}

\caption{The system prompt for the VLM (Doubao and Gemini) to assess \textbf{doodle editing}.}
    \label{tab:eva_edit3}
\end{minipage}
\end{table*}

\begin{table*}[h!]\centering

\begin{minipage}{1\linewidth}\vspace{0mm}    \centering
\begin{tcolorbox} 
    \small
You are a professional image editing evaluator. I will provide you with a source image (which may contain collaged cropped images), a target image, and an instruction (which may or may not be present). Your task is to evaluate the target image based on the following criteria:

Consistency of Non-Edited Regions: Assess whether the areas or objects in the source image that were not edited (according to the instructions and textures) remain consistent with the target image. Specifically, compare whether the appearance and position of each object in the non-edited areas of the source and target images are exactly the same, and evaluate whether there are any significant color discrepancies, such as yellowing, whitening, sharpening, or blurring.

Image Fusion and Instruction-based Editing Accuracy: You need to determine if the model has successfully integrated the textures with the surrounding environment. Additionally, if the user has provided instructions, evaluate whether the textures in the target image accurately merge with the surroundings in the manner specified by the user's instructions.

Quality of the Editing Result: Evaluate whether the edited objects blend naturally into the target image. The naturalness is primarily assessed based on factors such as lighting, shadows, and the size proportions of the objects. Furthermore, check whether the edited objects make it obvious to the user that they are fake or unnatural.

Output the result in JSON format as: {"judge": True/False, "reason": "xxx"}. Please evaluate the target image step by step based on the three criteria. Only return True if all criteria are satisfied. Provide reasoning for each of the three points.

\end{tcolorbox}

\caption{The system prompt for the VLM (Doubao and Gemini) to assess \textbf{image fusion}.}
    \label{tab:eva_edit4}
\end{minipage}
\end{table*}

\begin{table*}[h!]\centering
\begin{minipage}{1\linewidth}\vspace{0mm}    \centering
\begin{tcolorbox} 
    \small

You are a professional reference image generation evaluator. I will provide you with reference images  (which may contain scribbles), the final target image, and an instruction for generating the target image based on the reference image. Your task is to evaluate the target image based on the following criteria:

1. Consistency: Check whether the elements requested in the instruction, which are derived from the reference image (such as people, objects, or abstract attributes), are consistent with the corresponding elements in the target image.
2. Scribbles and Instruction Adherence: Verify whether the target image has been generated accurately according to the given instruction. Ensure that the changes and features requested in the instruction are correctly implemented in the target image. Please note that you must specifically assess whether the generated object is located in the position of the scribble as specified in the instructions.
3. Visual Integrity: Assess whether there are any noticeable issues of visual breakdown or distortion in the target image (such as unrealistic blending, unnatural proportions, or mismatched lighting and shadows).
4. Generation Quality: Evaluate whether the reference objects or attributes have been directly copied and pasted, which could lead to inconsistencies such as missing parts, or unnatural duplication, or result in elements in the target image that are too similar to the reference, making the final image feel forced or unnatural.
5. Check for Scribbles in the Output: Check whether the scribbles input by the user are also output in the target image. It is not acceptable for the scribbles to be included in the output image.

Output the result in JSON format as: {"judge": True/False, "reason": "xxx"}. Please evaluate the target image step by step based on the five criteria. Only return True if all criteria are satisfied. Provide reasoning for each of the five points.

\end{tcolorbox}

\caption{The system prompt for the VLM (Doubao and Gemini) to assess \textbf{scribble and multimodal instruction-based generation}.}
    \label{tab:eva_gen1}
\end{minipage}
\end{table*}

\begin{table*}[h!]\centering
\begin{minipage}{1\linewidth}\vspace{0mm}    \centering
\begin{tcolorbox} 
    \small

You are a professional reference image generation evaluator. I will provide you a reference image  (which contains scribbles), the final target image, and an instruction for generating the target image based on the reference image. Your task is to evaluate the target image based on the following criteria:

1. Scribbles and Instruction Adherence: Verify whether the target image has been generated accurately according to the given instruction and reference image. You must specifically assess whether the generated object is located in the position of the scribble as specified in the reference image and instruction.
2. Visual Integrity: Assess whether there are any noticeable issues of visual breakdown or distortion in the target image (such as unrealistic blending, unnatural proportions, or mismatched lighting and shadows).
3. Generation Quality: Evaluate whether the reference objects or attributes have been directly copied and pasted, which could lead to inconsistencies such as missing parts, or unnatural duplication, or result in elements in the target image that are too similar to the reference, making the final image feel forced or unnatural.
4. Check for Scribbles in the Output: Check whether the scribbles input by the user are also output in the target image. It is not acceptable for the scribbles to be included in the output image.

Output the result in JSON format as: {"judge": True/False, "reason": "xxx"}. Please evaluate the target image step by step based on the four criteria. Only return True if all criteria are satisfied. Provide reasoning for each of the four points.

\end{tcolorbox}

\caption{The system prompt for the VLM (Doubao and Gemini) to assess \textbf{scribble instruction-based generation}.}
    \label{tab:eva_gen2}
\end{minipage}
\end{table*}

\begin{table*}[h!]\centering
\begin{minipage}{1\linewidth}\vspace{0mm}    \centering
\begin{tcolorbox} 
    \small

You are a professional reference image generation evaluator. I will provide you a reference image  (which contains scribbles), the final target image, and an instruction (which may or may not be present). for generating the target image based on the reference image. Your task is to evaluate the target image based on the following criteria:

1. Scribbles and Instruction Adherence: Verify whether the target image has been generated accurately according to the given instruction and scribbles in reference image. You must specifically assess whether the generated object is located in the position of the scribble as specified in the reference image and instruction. The generation result should resemble the scribble but be transformed into a more aesthetically pleasing actual object, rather than just the scribble lines. Additionally, if the user provided instructions, evaluate
whether the generated result meets the requirements of the instructions.
2. Visual Integrity: Assess whether there are any noticeable issues of visual breakdown or distortion in the target image (such as unrealistic blending, unnatural proportions, or mismatched lighting and shadows).
3. Generation Quality: Evaluate whether the reference objects or attributes have been directly copied and pasted, which could lead to inconsistencies such as missing parts, or unnatural duplication, or result in elements in the target image that are too similar to the reference, making the final image feel forced or unnatural.
4. Check for Scribbles in the Output: Check whether the scribbles input by the user are also output in the target image. It is not acceptable for the scribbles to be included in the output image.

Output the result in JSON format as: {"judge": True/False, "reason": "xxx"}. Please evaluate the target image step by step based on the four criteria. Only return True if all criteria are satisfied. Provide reasoning for each of the four points.

\end{tcolorbox}

\caption{The system prompt for the VLM (Doubao and Gemini) to assess \textbf{doodle generation}.}
    \label{tab:eva_gen3}
\end{minipage}
\end{table*}

\section{DreamOmni3 Benchmark}
\label{sec:benchmark}
\begin{figure*}[h]
\centering
    \captionsetup{type=figure}
    \includegraphics[width=1\linewidth]{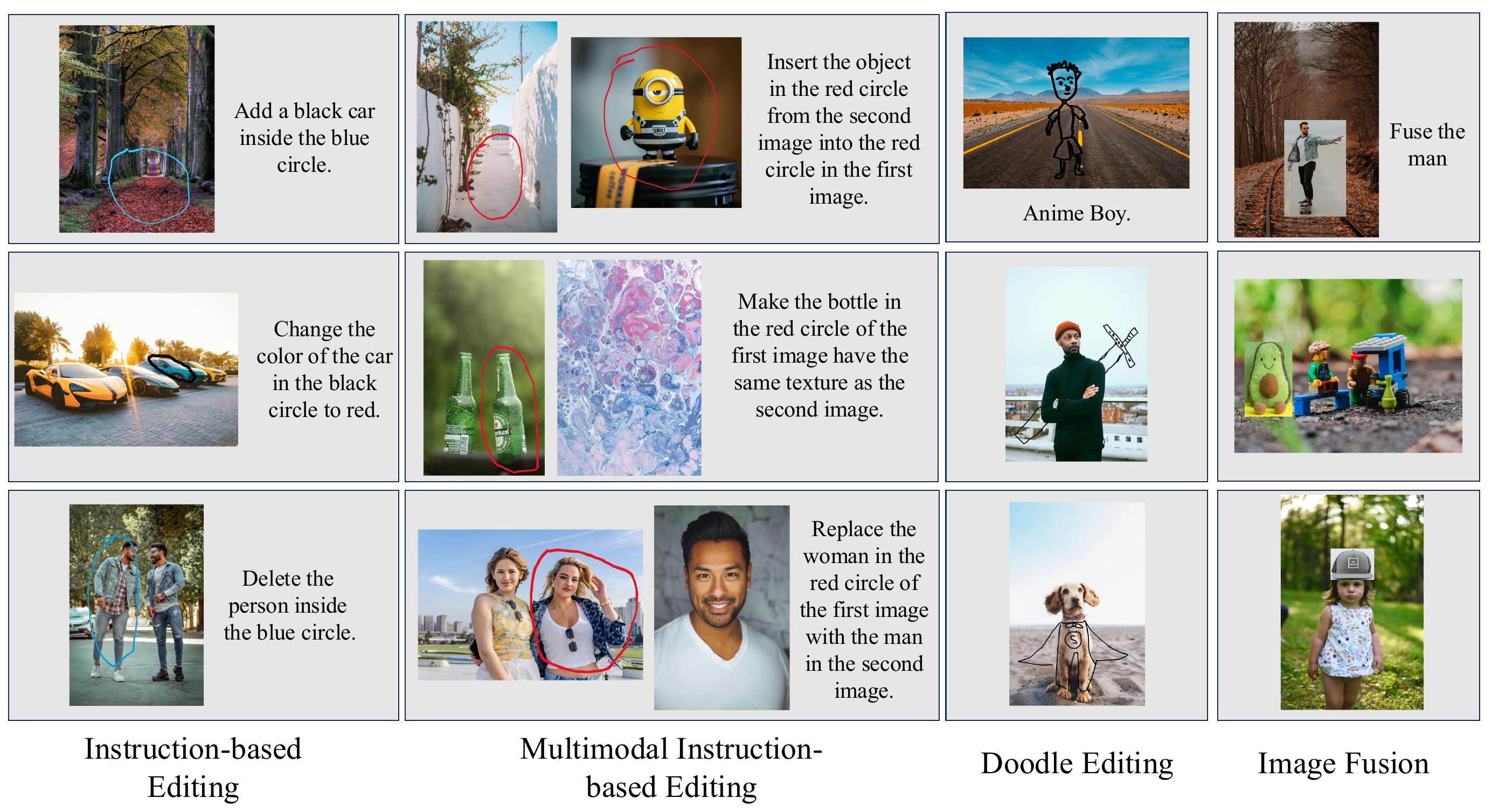}
    \captionof{figure}{Examples of scribble-based editing in DreamOmni3 benchmark.  }
    \label{fig:sup-benchmark-edit}
    
\end{figure*}

\begin{figure*}[h]
\centering
    \captionsetup{type=figure}
    \includegraphics[width=1\linewidth]{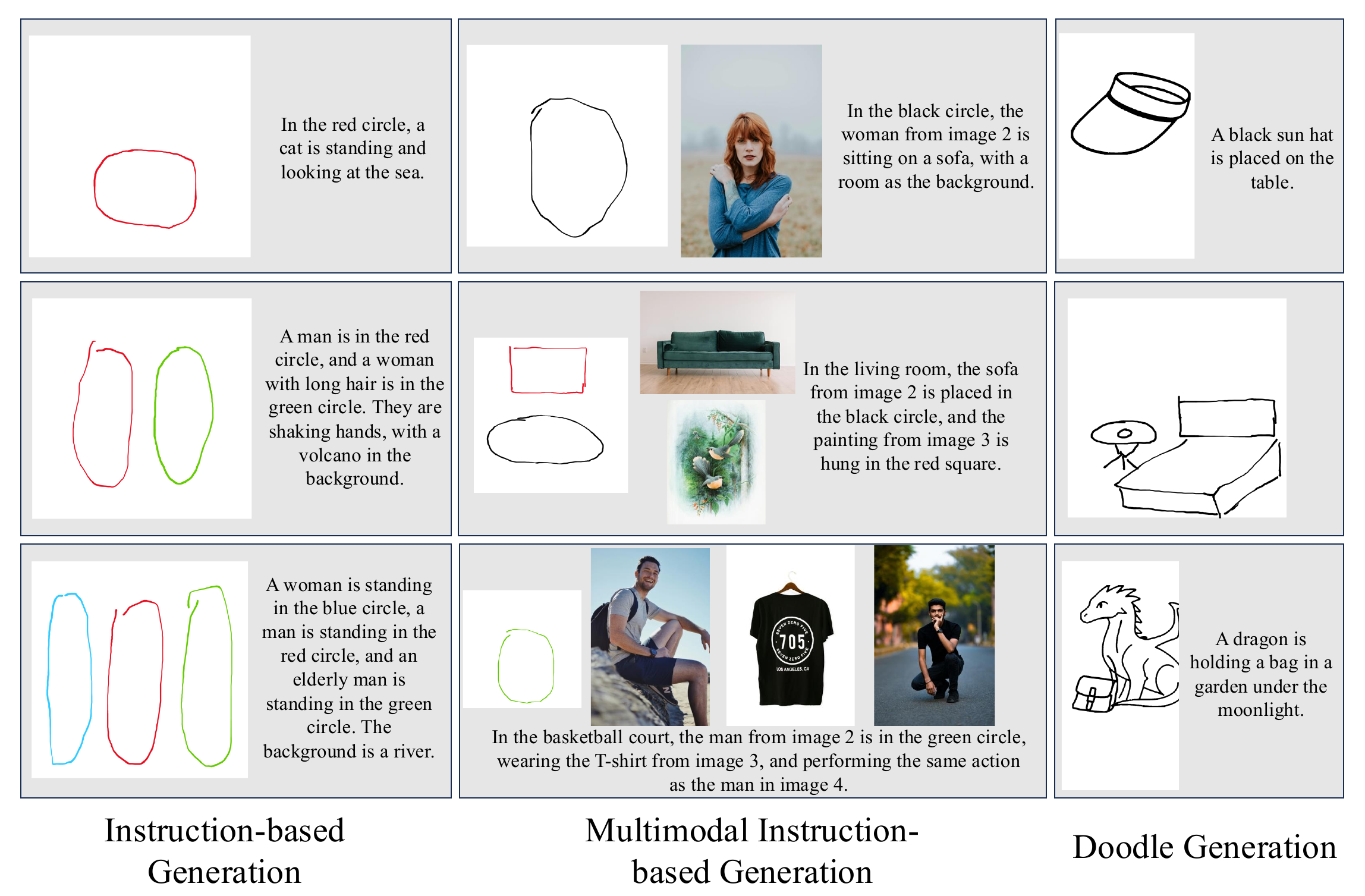}
    \captionof{figure}{Examples of scribble-based generation in DreamOmni3 benchmark. }
    \label{fig:sup-benchmark-gen}
    
\end{figure*}

The DreamOmni3 benchmark includes $136$ scribble-based editing test cases and $88$ scribble-based generation test cases. Visualizations of the editing and generation test cases are shown in Fig.~\ref{fig:sup-benchmark-edit} and Fig.~\ref{fig:sup-benchmark-gen}, respectively. The benchmark covers a wide range of test cases, with input reference images ranging from one to four. The DreamOmni3 benchmark includes four types of scribble-based editing (instruction-based editing, multimodal instruction-based editing, doodle editing, and image fusion) and three types of scribble-based generation (instruction-based generation, multimodal instruction-based generation, and doodle generation). Our editing benchmark thoroughly integrates language, images, and scribbles as instructions for editing and generation, with the design objects including a variety of concrete objects and abstract attributes, enabling comprehensive evaluation of performance in different scenarios. The DreamOmni3 Benchmark will be released.

\section{More Scribble-based Editing Cases}
\label{sec:more_editing}
\begin{figure*}[h]
\centering
    \captionsetup{type=figure}
    \includegraphics[width=.8\linewidth]{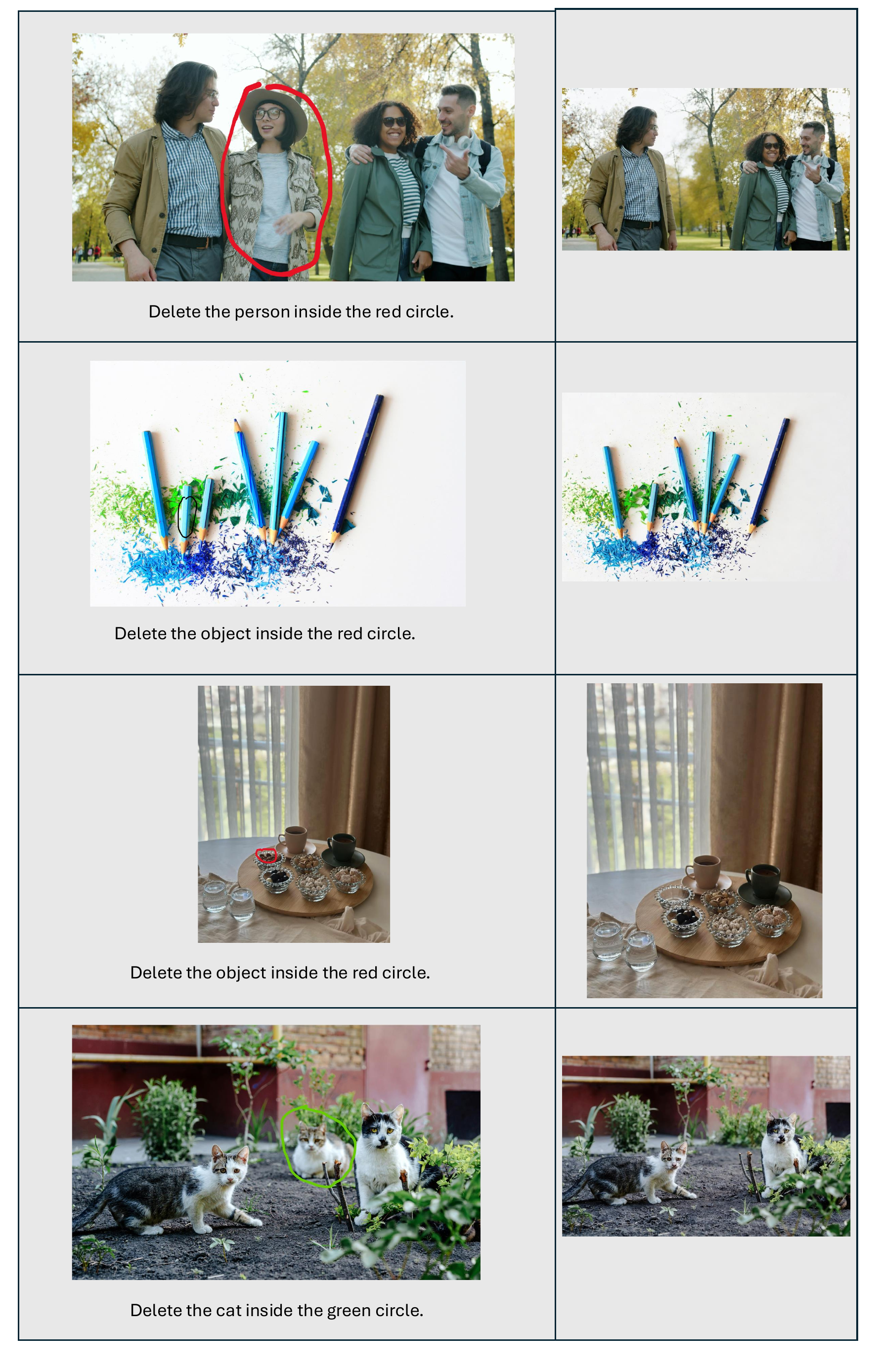}
    \captionof{figure}{Scribble-based editing cases of DreamOmni3.  }
    \label{fig:sup-edit-show1}
\end{figure*}

\begin{figure*}[h]
\centering
    \captionsetup{type=figure}
    \includegraphics[width=.8\linewidth]{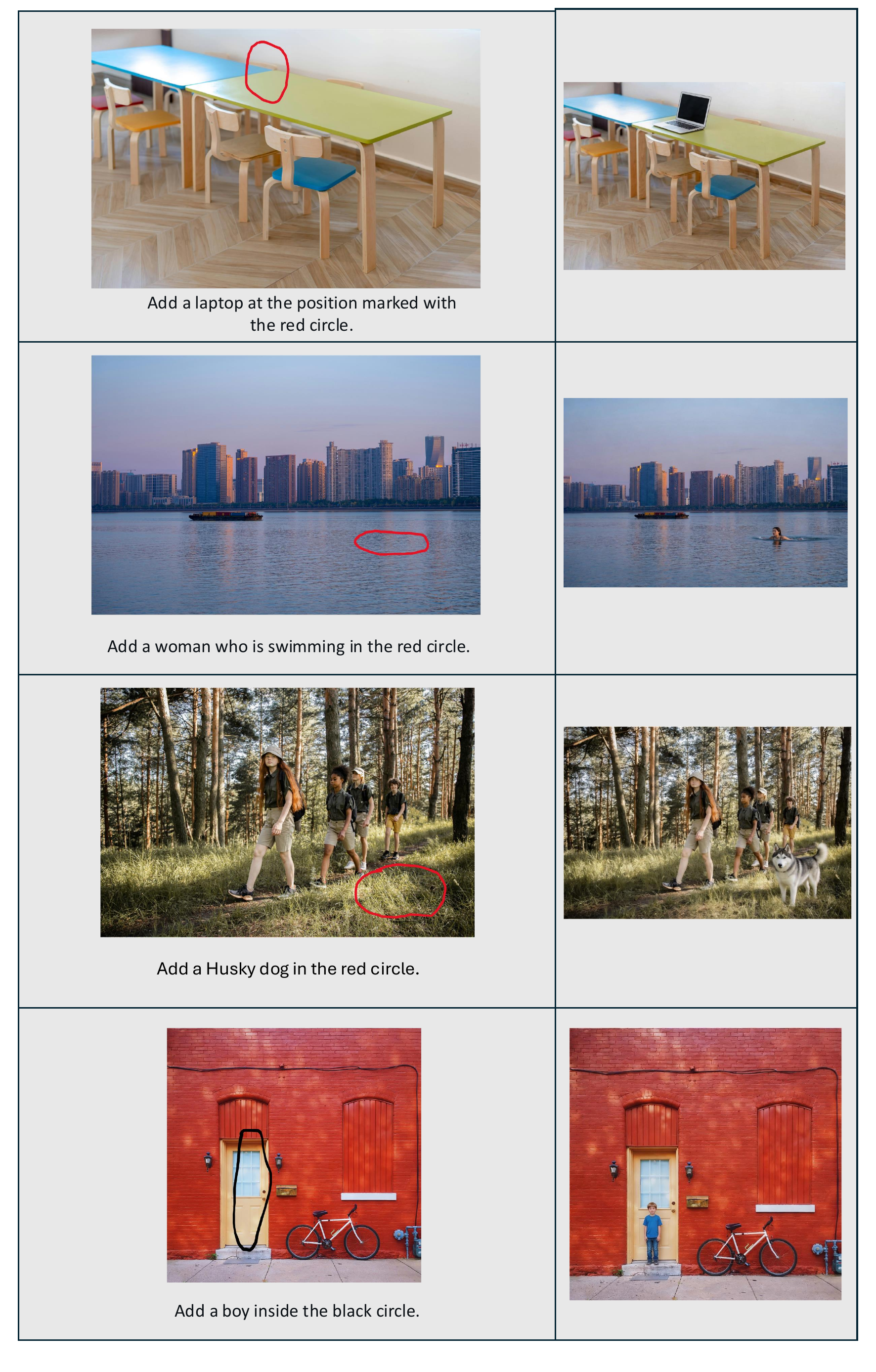}
    \captionof{figure}{Scribble-based editing cases of DreamOmni3.  }
    \label{fig:sup-edit-show2}
\end{figure*}

\begin{figure*}[h]
\centering
    \captionsetup{type=figure}
    \includegraphics[width=.8\linewidth]{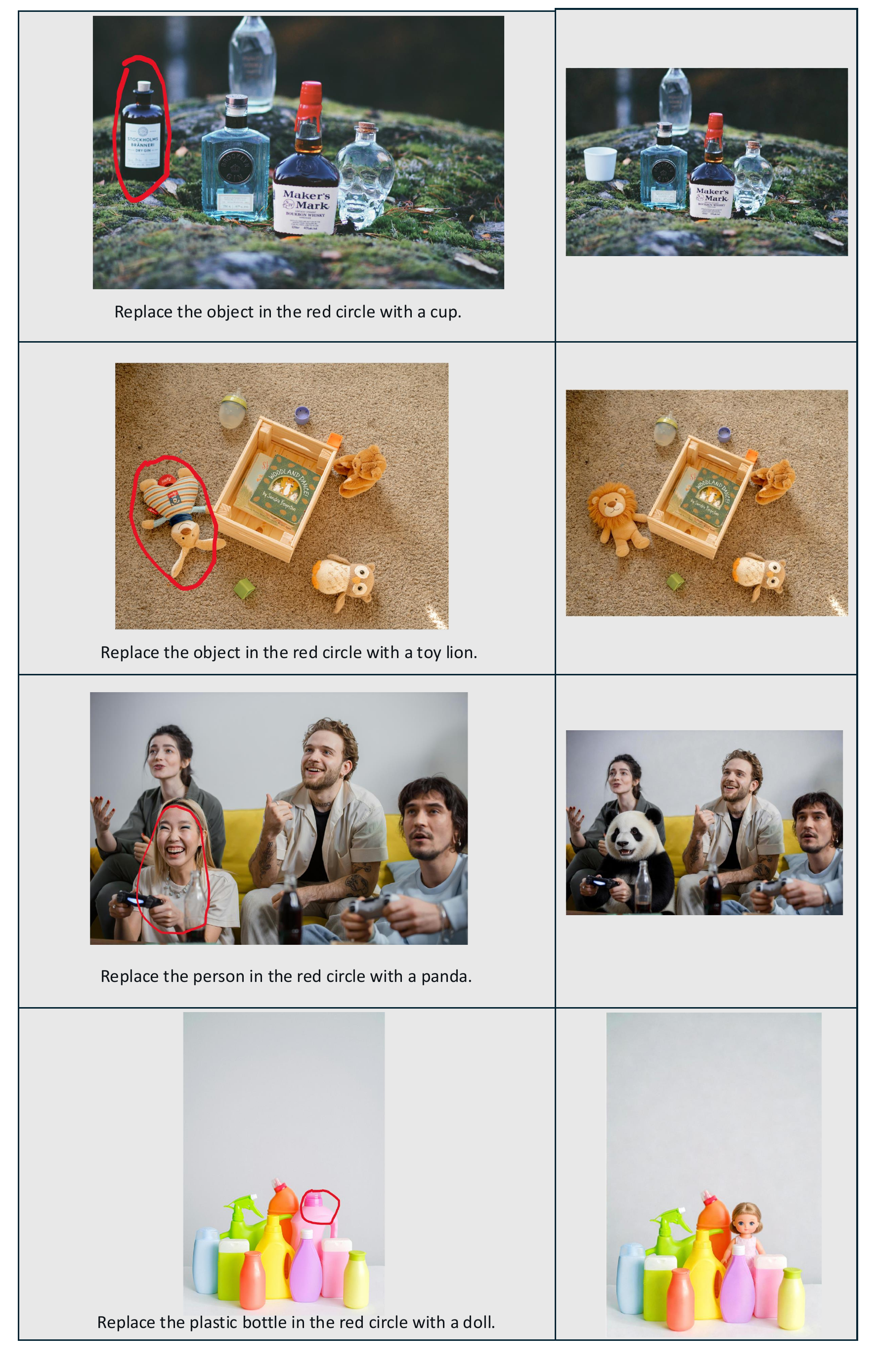}
    \captionof{figure}{Scribble-based editing cases of DreamOmni3.  }
    \label{fig:sup-edit-show3}
\end{figure*}

\begin{figure*}[h]
\centering
    \captionsetup{type=figure}
    \includegraphics[width=.8\linewidth]{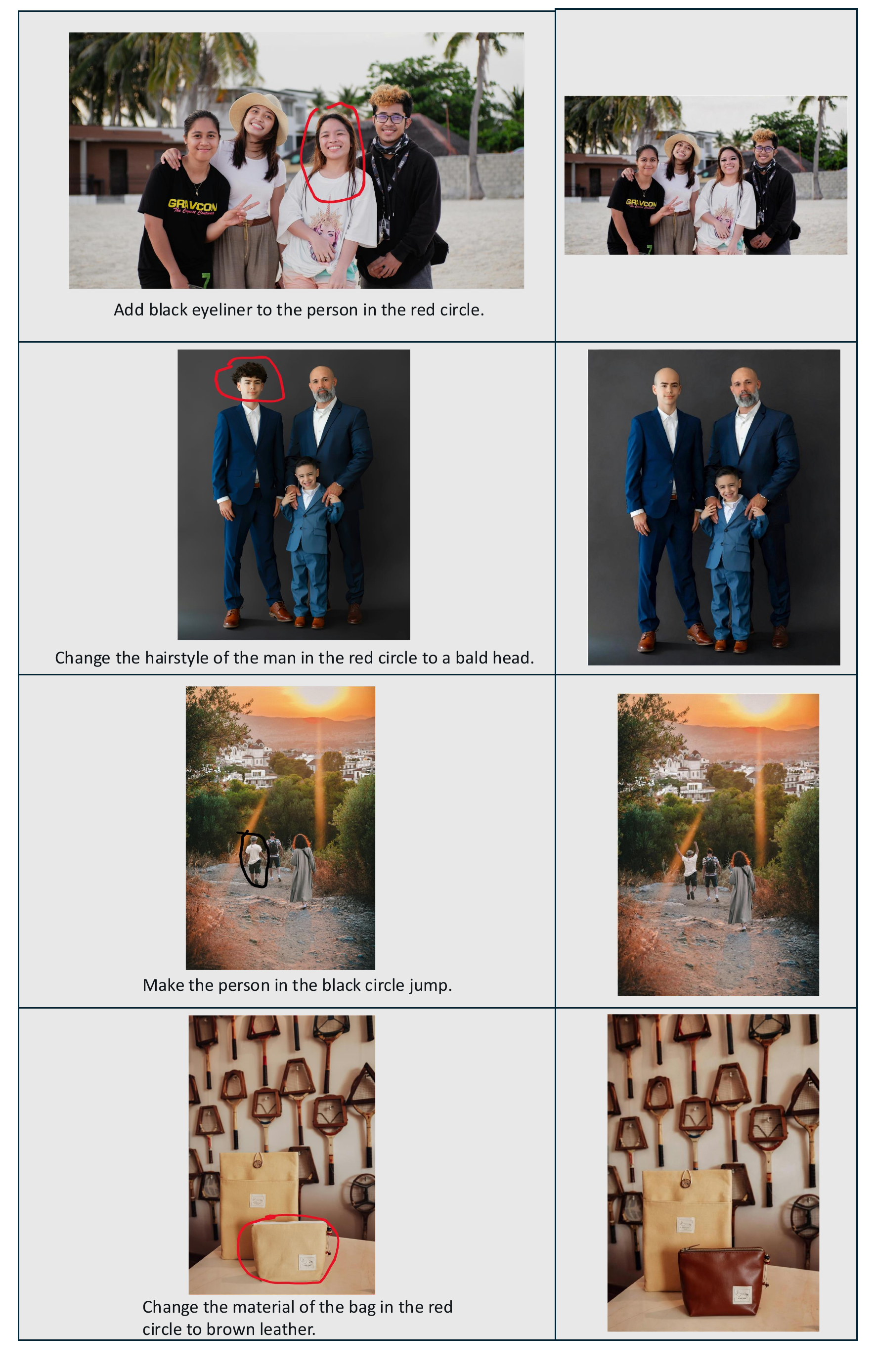}
    \captionof{figure}{Scribble-based editing cases of DreamOmni3.  }
    \label{fig:sup-edit-show4}
\end{figure*}

\begin{figure*}[h]
\centering
    \captionsetup{type=figure}
    \includegraphics[width=.8\linewidth]{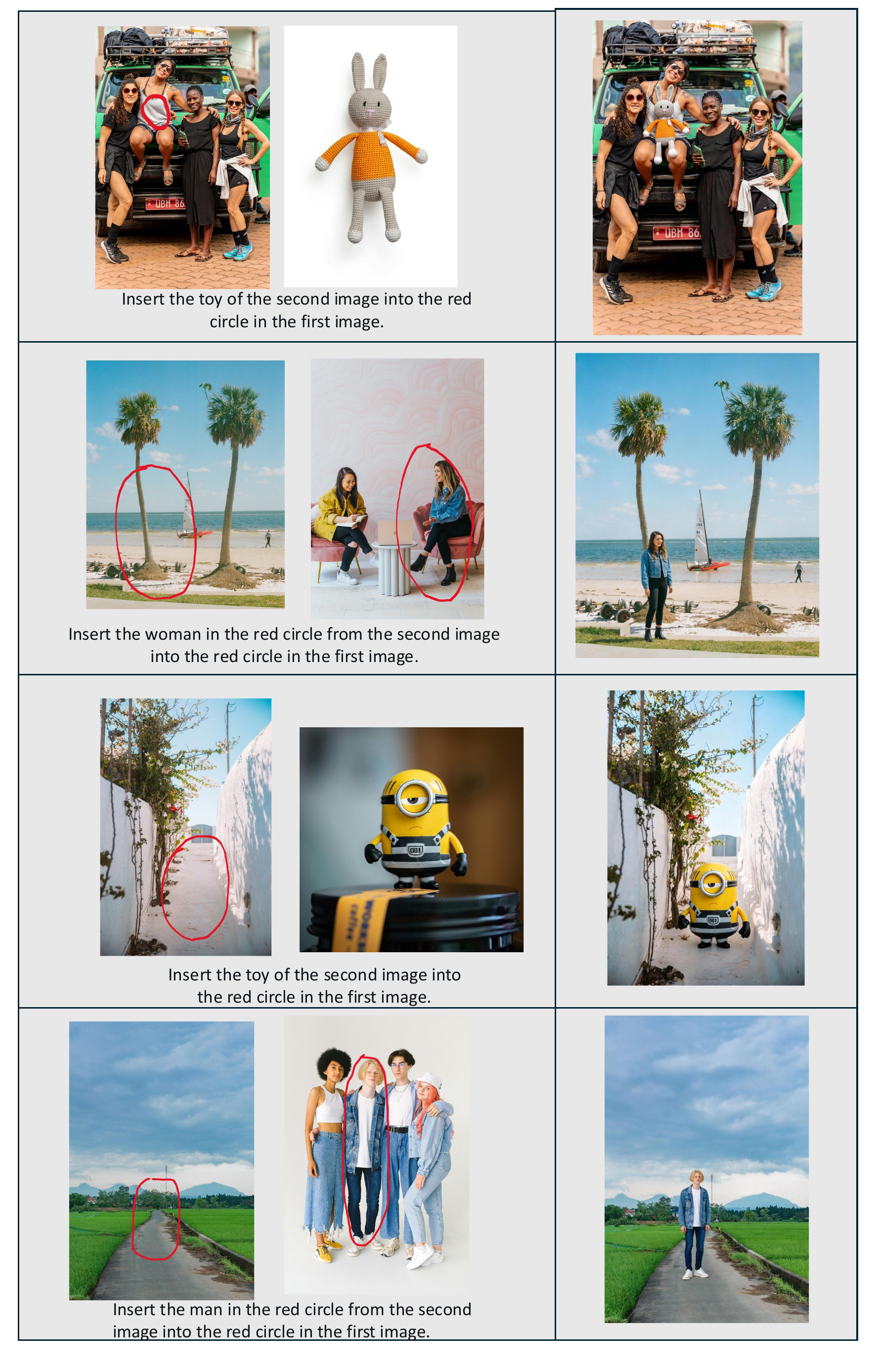}
    \captionof{figure}{Scribble-based editing cases of DreamOmni3.  }
    \label{fig:sup-edit-show5}
\end{figure*}

\begin{figure*}[h]
\centering
    \captionsetup{type=figure}
    \includegraphics[width=.8\linewidth]{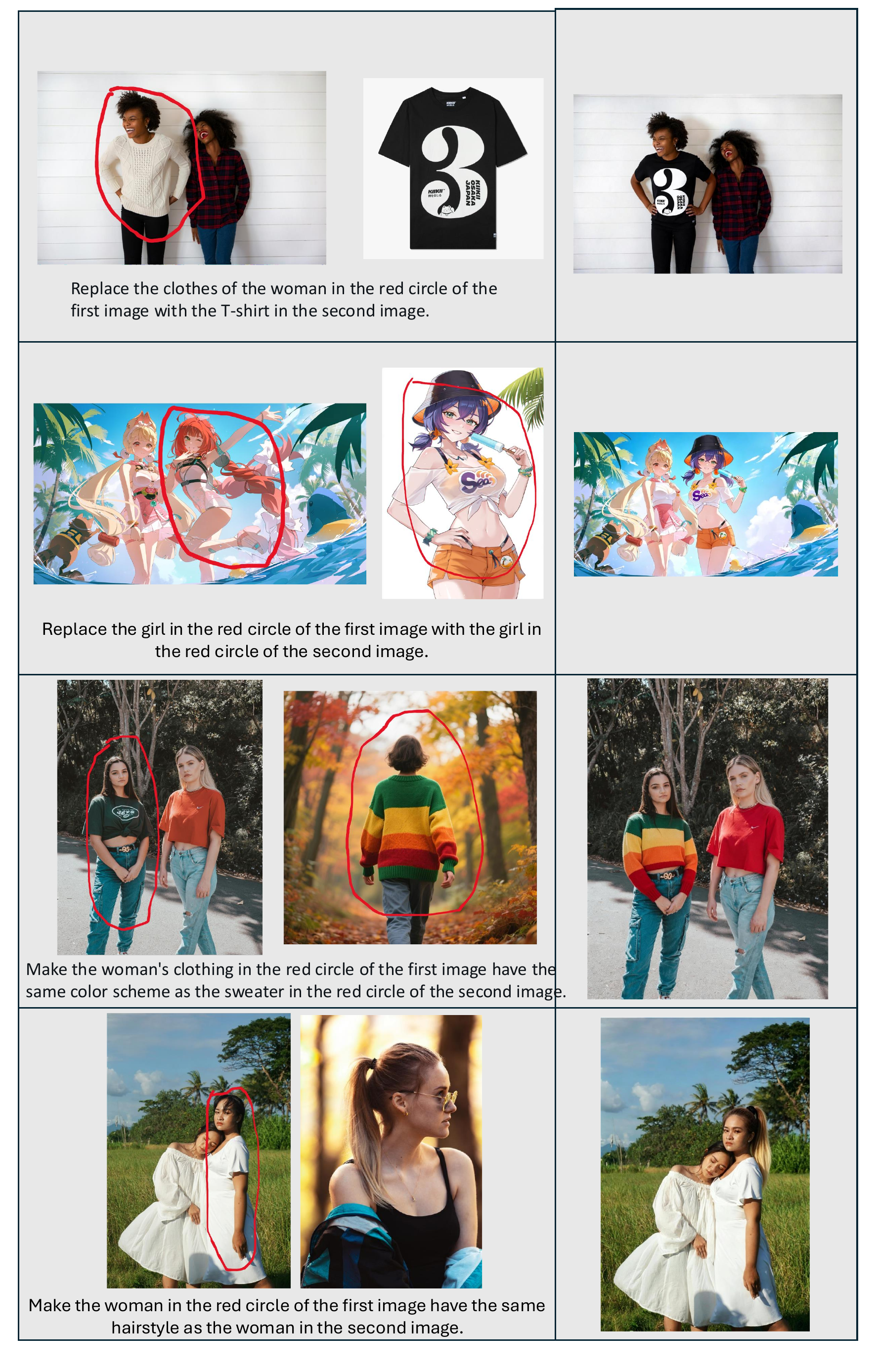}
    \captionof{figure}{Scribble-based editing cases of DreamOmni3.  }
    \label{fig:sup-edit-show6}
\end{figure*}

\begin{figure*}[h]
\centering
    \captionsetup{type=figure}
    \includegraphics[width=.8\linewidth]{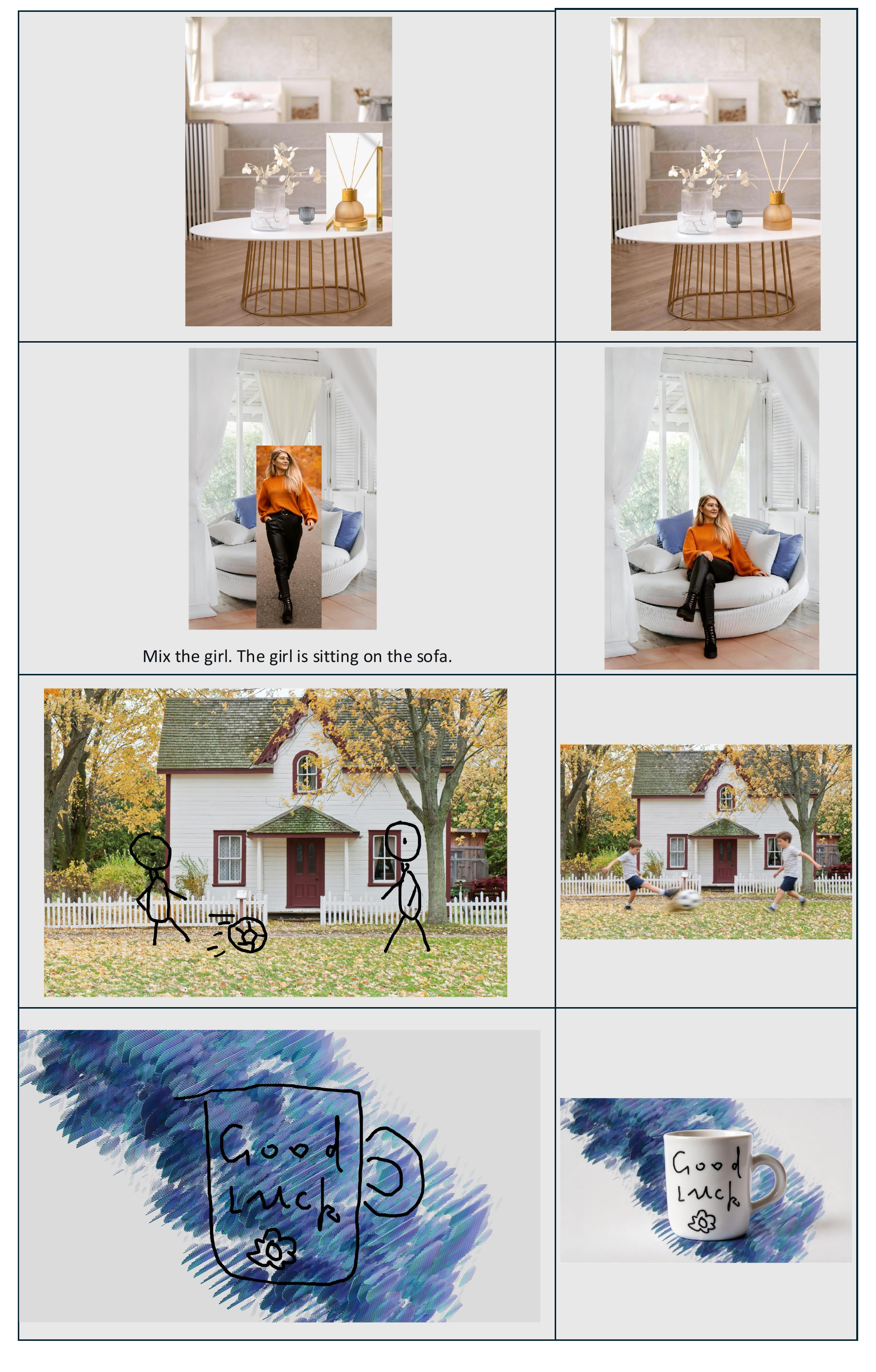}
    \captionof{figure}{Scribble-based editing cases of DreamOmni3.  }
    \label{fig:sup-edit-show7}
\end{figure*}

As shown in Fig.~\ref{fig:sup-edit-show1}, Fig.~\ref{fig:sup-edit-show2}, Fig.~\ref{fig:sup-edit-show3}, Fig.~\ref{fig:sup-edit-show4}, Fig.~\ref{fig:sup-edit-show5}, Fig.~\ref{fig:sup-edit-show6}, and Fig.~\ref{fig:sup-edit-show7}, we present additional visual cases of DreamOmni3 on the scribble-based editing task.

\section{More Scribble-based Generation Cases}
\label{sec:more_gen}
\begin{figure*}[h]
\centering
    \captionsetup{type=figure}
    \includegraphics[width=.8\linewidth]{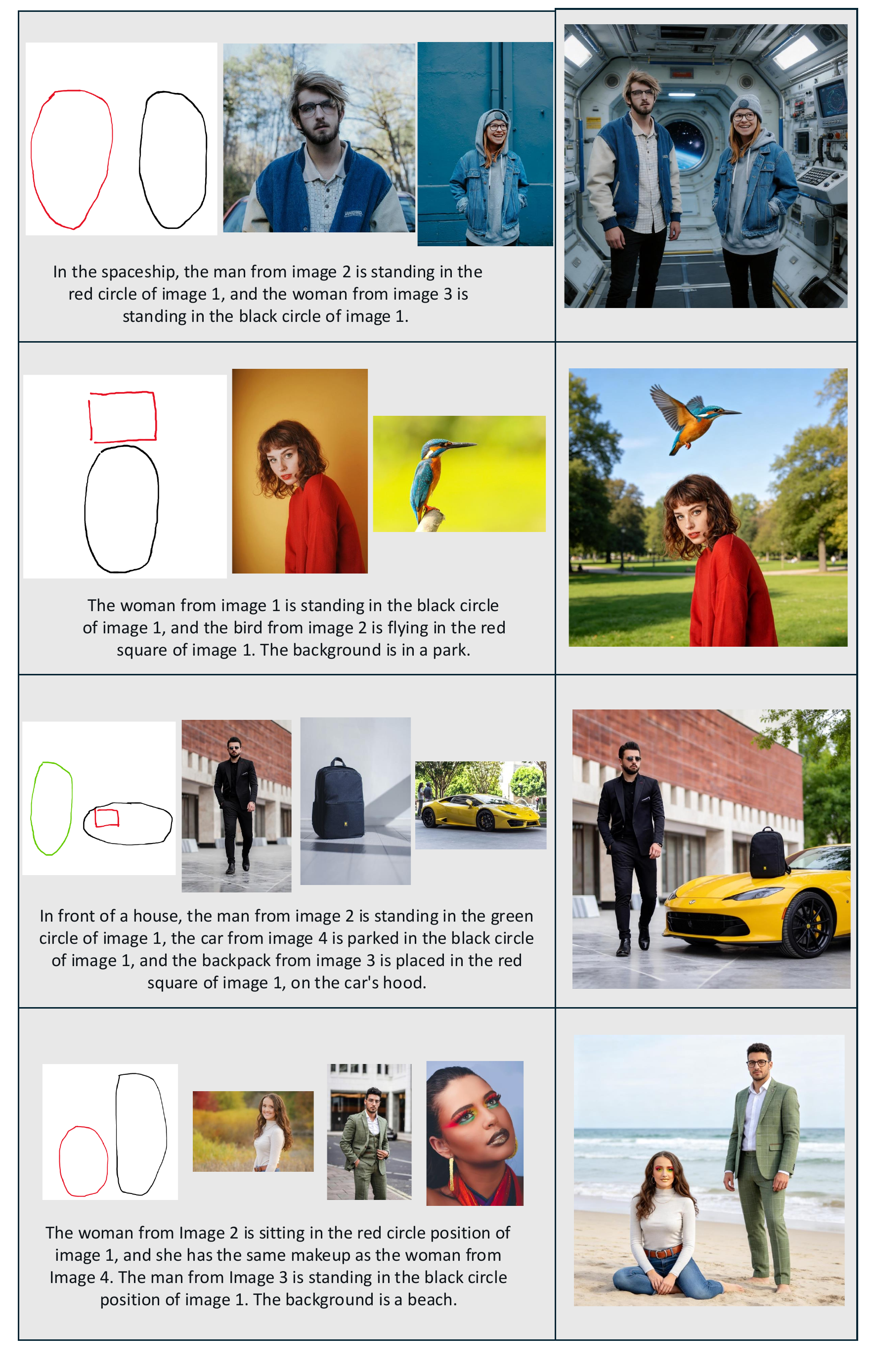}
    \vspace{-6mm}
    \captionof{figure}{Scribble-based generation cases of DreamOmni3.  }
    \label{fig:sup-gen-show1}
    \vspace{-5mm}
\end{figure*}

\begin{figure*}[h]
\centering
    \captionsetup{type=figure}
    \includegraphics[width=.8\linewidth]{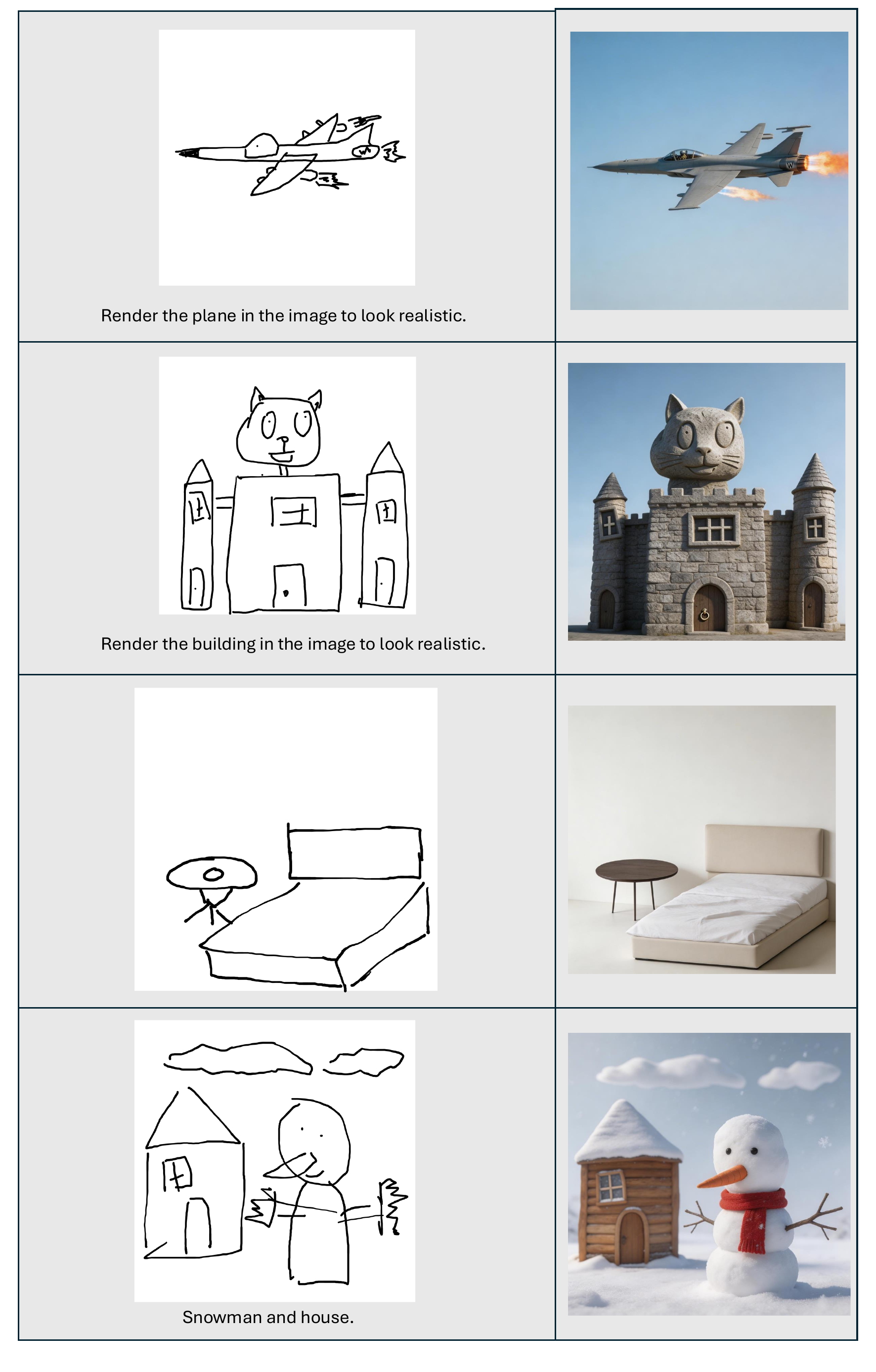}
    \vspace{-6mm}
    \captionof{figure}{Scribble-based generation cases of DreamOmni3.  }
    \label{fig:sup-gen-show2}
    \vspace{-5mm}
\end{figure*}

As shown in Fig.~\ref{fig:sup-gen-show1} and Fig.~\ref{fig:sup-gen-show2}, we present additional visual cases of DreamOmni3 on the scribble-based generation task.

\end{document}

%% file: macros.tex
\usepackage{graphicx}
\usepackage{amssymb}
\usepackage{pifont}
\usepackage{floatrow}
\usepackage{amsmath} 
\usepackage{float}
\usepackage{wrapfig}
\usepackage{multirow}
\usepackage{tcolorbox}
\tcbuselibrary{breakable, skins, raster}
\usepackage{listings}